\documentclass[11pt,a4paper]{article}
\usepackage{times,latexsym}
\usepackage{url}
\usepackage[T1]{fontenc}
\usepackage[font={small}]{caption}
\usepackage[symbol]{footmisc}
\usepackage{listings}
\usepackage{enumitem}

\usepackage[acceptedWithA]{tacl2021v1}

\usepackage{xspace,mfirstuc,tabulary}

\usepackage{float} 

\usepackage{graphicx}
\usepackage{multirow}
\usepackage{booktabs}
\usepackage{amsfonts}       
\usepackage{amsmath}        
\usepackage{algorithm}      
\usepackage{algpseudocode}  
\usepackage{listings}       
\usepackage{enumitem}       
\usepackage{xcolor}         
\usepackage{subcaption}     
\usepackage{babel}
\usepackage{natbib}
\usepackage{stmaryrd}       
\usepackage{wrapfig}
\usepackage{moresize}
\newif\iftaclinstructions
\taclinstructionsfalse 
\iftaclinstructions
\renewcommand{\confidential}{}
\renewcommand{\anonsubtext}{(No author info supplied here, for consistency with
TACL-submission anonymization requirements)}
\newcommand{\instr}
\fi

\iftaclpubformat 

\else

\fi

\title{Understanding Epistemic Language with a \\ Language-augmented Bayesian Theory of Mind}

\author{
  Lance Ying$^{1,2*}$,
  Tan Zhi-Xuan$^{1*}$, 
  Lionel Wong$^1$ \\
  {\bf
  Vikash Mansinghka$^1$,
  Joshua B. Tenenbaum$^1$
  } \\
  $^1$Massachusetts Institute of Technology, Cambridge, MA\\
  $^2$Harvard University, Cambridge, MA 
}

\date{}

\begin{document}
\maketitle
\def\thefootnote{*}\footnotetext{Equal Contribution.}\def\thefootnote{\arabic{footnote}}
\begin{abstract}
How do people understand and evaluate claims about others' beliefs, even though these beliefs cannot be directly observed? In this paper, we introduce a cognitive model of epistemic language interpretation, grounded in Bayesian inferences about other agents' goals, beliefs, and intentions: a language-augmented Bayesian theory-of-mind (LaBToM). By translating natural language into an epistemic ``language-of-thought'' with grammar-constrained LLM decoding, then evaluating these translations against the inferences produced by inverting a generative model of rational action and perception, LaBToM captures graded plausibility judgments of epistemic claims. We validate our model in an experiment where participants watch an agent navigate a maze to find keys hidden in boxes needed to reach their goal, then rate sentences about the agent's beliefs. In contrast with multimodal LLMs (GPT-4o, Gemini Pro) and ablated models, our model correlates highly with human judgments for a wide range of expressions, including modal language, uncertainty expressions, knowledge claims, likelihood comparisons, and attributions of false belief.
\end{abstract}

\begin{figure*}[t!]
    \centering
    \includegraphics[width=0.915\textwidth]{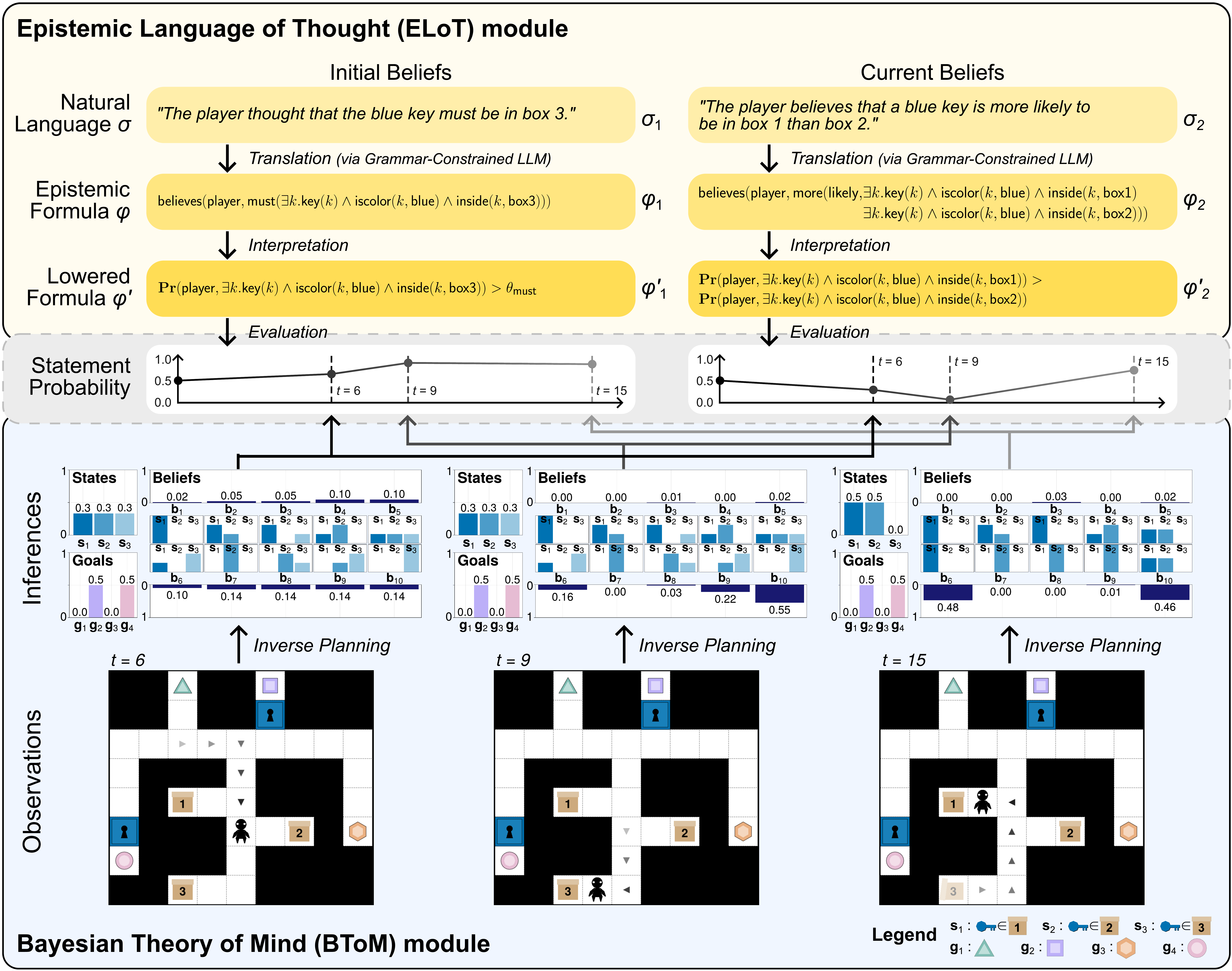}
    \caption{\textbf{Overview of our model, a Language-augmented Bayesian Theory of Mind (LaBToM)}. Here, our model evaluates the plausibility of epistemic language about a player character trying to find keys in boxes so as to reach one of four goals (gems with different shapes/colors, $\textsf{g}_1$-- $\textsf{g}_4$) which may be locked behind doors. \textbf{(Top)} We translate natural language statements about the player's initial ($\sigma_1$) and current beliefs ($\sigma_2$) into an unambiguous \emph{epistemic language of thought (ELoT)} via grammar-constrained LLM parsing. ELoT statements (i.e. epistemic formulas $\varphi$) are interpreted with a probability-based semantics (lowered formulas $\varphi'$). \textbf{(Bottom)} We use our \emph{Bayesian theory-of-mind (BToM)} module to produce inferences (bar charts) about the environment state (top left), the player's goal (bottom left), and the player's belief state (right) at each step $t$, given observations across time. Each possible belief state $\textsf{b}_i$ is itself a distribution over environment states $\textsf{s}_j$, where state $\textsf{s}_j$ corresponds to a blue key being in box $j$. \textbf{(Middle)} We evaluate the posterior probability$^*$ of each ELoT statement $\varphi$ from the BToM inferences at each step $t$ ($^*$under a 50-50 prior over statement truth). Statement $\sigma_1$ (\emph{\footnotesize``The player thought that the blue key must be in box 3.''}) increases in probability from $t$$=$$6$ to $t$$=$$9$, then stays high, since it becomes clear from the player looking in box 3 at $t$$=$$9$ that they \emph{initially} thought a key must be in box 3. Statement $\sigma_2$ (\emph{\footnotesize``The player believes that a blue key is more likely to be in box 1 than box 2.''}) decreases in probability as the player walks away from box 1 at $t$$=$$6$ through $t$$=$$9$. However, when the player finds box 3 empty, then moves past box 2 up to box 1 at $t$$=$$15$, it becomes much more probable that the player \emph{currently} thinks a key is more likely to be in box 1 rather than box 2.}
    \label{fig:overview}
    \vspace{-9pt}
\end{figure*}

\section{Introduction}
\label{sec:introduction}
People regularly use and interpret language about other agents' beliefs, evaluating rich linguistic constructions that may involve claims about what others' consider possible (\emph{``Grace thinks Katie might have eaten the cookie''}), what they find more probable (\emph{``Tom thinks Sam is the most likely to win the election''}), or the relationship of their beliefs to the world (\emph{``John didn't know that today was a holiday''}).  But since these beliefs are not directly observable, how do people judge the truth or acceptability of epistemic claims? Philosophers and linguists have long investigated the semantics of epistemic language \cite{hintakka1962knowledge,partee1973semantics,loar1981mind}, offering compositional theories of how epistemic claims relate to the sets of worlds deemed possible by an agent \cite{von2011intensional}.
However, these theories do not explain how people ground epistemic language in its context of utterance. If someone says \emph{``Alice believes it might rain''}, how does her behavior (e.g. bringing an umbrella) render that statement more or less plausible or acceptable?

In this paper, we present a computational model of how people interpret epistemic language in context, grounded in inferences about what other agents believe (Figure \ref{fig:overview}) --- a \emph{Language-augmented Bayesian theory-of-mind} (LaBToM).
We build upon \emph{Bayesian theory-of-mind} (BToM), a framework which casts human mentalizing as probabilistic inference over a generative model of rational action and perception \cite{baker2017rational,jara2020naive}. We combine this framework with the compositionality afforded by probabilistic extensions of the language-of-thought hypothesis \cite{piantadosi2011learning,goodman2015probabilistic}, developing an \emph{epistemic language of thought} (ELoT) to represent how others represent the world. We parse natural language to this representation using grammar-constrained sequential Monte Carlo decoding \cite{loula2025syntactic} of large language models (LLMs), defining a flexible but structured probabilistic mapping from epistemic language to epistemic concepts \cite{wong2023word}. This allows us to quantify the plausibility of epistemic sentences against BToM inferences.

To evaluate our model, we run an experiment where participants watch animations of a player solving a gridworld puzzle called Doors, Keys, \& Gems \cite{zhi2020online}. In these puzzles, the player has to pick up (single-use) keys that may be hidden in boxes, using them to unlock doors of the same color to reach one of four valuable gems. The player's beliefs and goals are unknown to our participants, so they must \emph{infer} these mental states. We ask one set of participants to write sentences describing the past and present beliefs of the player, collecting a rich dataset of epistemic language that includes modal verbs, uncertainty expressions, knowledge claims, and descriptions of false beliefs. We task another set of participants with evaluating these statements, asking them to rate how likely a statement is given the context. We find that the inferences produced by our LaBToM model correlate highly with these  human ratings. In contrast, using a lower-level formalization of epistemic claims leads to substantially more parsing errors than our ELoT representation, and ablated BToM models and multimodal LLM baselines fail to explain human sentence judgments. These findings illustrate the importance of a conceptual vocabulary that is aligned with the structure of epistemic language, and the need for a coherent theory-of-mind to systematically interpret such language.

\section{Related Work}
\label{sec:related-work}

\paragraph{Model-theoretic formal semantics of epistemic language.} Sentences about what agents believe or know have been studied extensively across philosophy and linguistics. Both \emph{epistemic modals} \citep{yalcin2007epistemic,coates1987epistemic,egan2011epistemic,macfarlane2011assessment}
and \emph{propositional attitudes} \citep{frege1948sense,russell1903principles,schiffer2003things} complicate compositional and model-theoretic approaches to assigning truth conditions. Standard treatments associate belief sentences with sets of compatible worlds \cite{hintakka1962knowledge,von2011intensional}, though the role of an agent's mental states in a truth-conditional semantics remains debated \citep{egan2011epistemic}. More recent work grounds the gradedness of belief claims in probability \cite{lassiter2017graded,moss2015semantics}, which we follow in this paper. Unlike purely model-theoretic approaches, however, our model relates epistemic sentences to a \emph{functional} theory of agents' mental states \citep{loar1981mind}, explaining how people ground these sentences in agent behavior.

\paragraph{Cognitive approaches to (epistemic) language interpretation.} 
Our model draws on accounts of propositional attitudes \citep{a1981propositional} and word meaning that emphasize the role of mental states \cite{loar1981mind,block1987advertisement,lake2023word} and the interface between language and conceptual content, as in cognitive or psychosemantics \cite{fodor1987psychosemantics,lakoff1988cognitive,jackendoff2003precis} and functional role semantics \cite{harman1982conceptual}. We build most closely upon theories that ground linguistic meaning in a (probabilistic) language of thought \citep{fodor1975language,goodman2015probabilistic,wong2023word, zhang2023grounded}, and models 
that map sentences to grounded symbolic representations of agents' behavior \citep{artzi2013weakly}.

\paragraph{Bayesian Theory-of-Mind.} To relate epistemic sentences to (inferred) mental states, we build on the Bayesian Theory-of-Mind framework \cite{baker2017rational,zhixuan2022solving}, and related work on epistemic action understanding \cite{croom2023seeing,shvo2020epistemic}. BToM provides a functional theory of propositional attitudes like knowledge and belief, leveraging the connection between symbolic representations of the world (used in (inverse) planning \cite{mcdermott1998pddl}) and our representations of others' minds.

\paragraph{Theory-of-Mind in Language Models.} With recent advances in LLM capabilities, some researchers have suggested that LLMs might serve as cognitive models \cite{binz2023turning}, including as models of theory-of-mind \cite{strachan2024testing}, thereby implicitly capturing the semantics of belief sentences \cite{piantadosi2022meaning}. However, while LLMs perform well on some ToM tasks, they do not reliably generalize \cite{shapira2024clever} to richer multi-step \cite{kim2023fantom} or multi-modal contexts \cite{jin2023mmtomqa, ying2024goma}. Our model instead uses (grammar-constrained) LLMs as a flexible mappings between language and formal meaning representations \cite{wong2023word,ying2023neuro,ying2023inferring,zhixuan2024pragmatic}, tying language to an \emph{explicit} semantics of epistemic concepts.

\section{Computational Model}
\label{sec:model}

Our LaBToM model comprises two interlinked modules. The first module (Figure \ref{fig:overview}, Top), an \emph{epistemic language of thought} (ELoT), models our capacity to compositionally represent the world (including the contents of others' minds) by combining more basic concepts into richer thoughts and expressions \cite{goodman2015probabilistic}, and how we flexibly translate such thoughts from natural language \cite{wong2023word}. The second module (Figure \ref{fig:overview}, Bottom), a \emph{Bayesian theory-of-mind} \cite{baker2017rational}, captures our intuitive inferences about others' minds via Bayesian inference over a generative model of how agents update their beliefs and act towards their goals. Epistemic language understanding can thus be modeled by mapping language into ELoT formulas, which we evaluate against the inferences produced by rational mentalizing.

\begin{table*}[t]
    \scriptsize
    \centering
    \begin{subtable}[b]{0.9\textwidth}
    \centering
    \begin{tabular}{@{}llll@{}}
    \toprule
    \textbf{Expression}                             & \textbf{Type}             & \textbf{Arg. Types}                                                          & \textbf{Definition}                                                                                               \\ \midrule
    \textbf{\emph{Belief Operators}}              &                           &                                                                              &                                                                                                                   \\
    $\textsf{believes}(A, \phi)$                    & $\mathcal{E}$             & $\mathcal{A}, \Phi$                                                          & $\textbf{Pr}(A, \phi) \geq \theta_{\textsf{believes}}$                                                            \\
    $\textsf{believes}_\textsf{modal}(A, M)$        & $\mathcal{E}$             & $\mathcal{A}, \mathcal{E} \slash \mathcal{A}$                                       & $M(A)$                                                                                                            \vspace{2pt} \\
    \textbf{\emph{Knowledge Operators}}           &                           &                                                                              &                                                                                                                   \\
    $\textsf{knows}_\textsf{that}(A, \phi)$         & $\mathcal{E}$             & $\mathcal{A}, \Phi$                                                          & $\textsf{believes}(A, \phi) \land \phi$                                                                           \\
    $\textsf{knows}_\textsf{if}(A, \phi)$           & $\mathcal{E}$             & $\mathcal{A}, \Phi$                                                          & $\textsf{knows}_\textsf{that}(A, \phi) \lor \textsf{knows}_\textsf{that}(A, \neg\phi)$                            \\
    $\textsf{knows}_\textsf{about}(A, C, \phi)$     & $\mathcal{E}$             & $\mathcal{A}, \Phi \slash \mathcal{O}, \Phi \slash \mathcal{O}$              & $\exists x. C(x) \land \textsf{knows}_\textsf{that}(A, \phi(x))$                                                  \\
    $\textsf{not-knows}_\textsf{that}(A, \phi)$        & $\mathcal{E}$             & $\mathcal{A}, \Phi$                                                          & $\neg\textsf{believes}(A, \phi) \land \phi$                                                                       \vspace{2pt} \\
    \textbf{\emph{Certainty Operators}}           &                           &                                                                              &                                                                                                                   \\
    $\textsf{certain}_\textsf{that}(A, \phi)$       & $\mathcal{E}$             & $\mathcal{A}, \Phi$                                                          & $\textbf{Pr}(A, \phi) \geq \theta_{\textsf{certain}}$                                                             \\
    $\textsf{certain}_\textsf{about}(A, C, \phi)$   & $\mathcal{E}$             & $\mathcal{A}, \Phi \slash \mathcal{O}, \Phi \slash \mathcal{O}$              & $\exists x. C(x) \land (\textbf{Pr}(A, \phi(x)) \geq \theta_{\textsf{certain}})$                                  \\
    $\textsf{uncertain}_\textsf{if}(A, \phi, \psi)$ & $\mathcal{E}$             & $\mathcal{A}, \Phi, \Phi$                                                    & $(\textbf{Pr}(A, \phi) < \theta_{\textsf{uncertain}}) \land (\textbf{Pr}(A, \psi) < \theta_{\textsf{uncertain}}$) \\
    $\textsf{uncertain}_\textsf{about}(A, C, \phi)$ & $\mathcal{E}$             & $\mathcal{A}, \Phi \slash \mathcal{O}, \Phi \slash \mathcal{O}$              & $\forall x. C(x) \to (\textbf{Pr}(A, \phi(x)) < \theta_{\textsf{uncertain}})$                                     \vspace{2pt} \\
    \textbf{\emph{Modal Verbs}}                   &                           &                                                                              &                                                                                                                   \\
    $\textsf{could}(\phi)$                          & $\mathcal{E} \slash \mathcal{A}$ & $\Phi$                                                                       & $\lambda A. \textbf{Pr}(A, \phi) \geq \theta_{\textsf{could}}$                                                    \\
    $\textsf{might}(\phi)$                          & $\mathcal{E} \slash \mathcal{A}$ & $\Phi$                                                                       & $\lambda A. \textbf{Pr}(A, \phi) \geq \theta_{\textsf{might}}$                                                    \\
    $\textsf{may}(\phi)$                            & $\mathcal{E} \slash \mathcal{A}$ & $\Phi$                                                                       & $\lambda A. \textbf{Pr}(A, \phi) \geq \theta_{\textsf{may}}$                                                      \\
    $\textsf{should}(\phi)$                         & $\mathcal{E} \slash \mathcal{A}$ & $\Phi$                                                                       & $\lambda A. \textbf{Pr}(A, \phi) \geq \theta_{\textsf{should}}$                                                   \\
    $\textsf{must}(\phi)$                           & $\mathcal{E} \slash \mathcal{A}$ & $\Phi$                                                                       & $\lambda A. \textbf{Pr}(A, \phi) \geq \theta_{\textsf{must}}$                                                     \vspace{2pt} \\
    \textbf{\emph{Modal Adjectives}}              &                           &                                                                              &                                                                                                                   \\
    $\textsf{likely}(\phi)$                         & $\mathcal{E} \slash \mathcal{A}$ & $\Phi$                                                                       & $\lambda A. \textbf{Pr}(A, \phi) \geq \theta_{\textsf{likely}}$                                                   \\
    $\textsf{degree}(\textsf{likely}, A, \phi)$     & $\Phi_\mathcal{F}$        & $\mathcal{A}, \Phi$                                                          & $\textbf{Pr}(A, \phi)$                                                                                           \vspace{2pt} \\
    \textbf{\emph{Comparatives}}                  &                           &                                                                              &                                                                                                                   \\
    $\textsf{more}(P, \phi, \psi)$                  & $\mathcal{E} \slash \mathcal{A}$ & $\mathcal{P}, \Phi, \Phi$                                                    & $\lambda A. \textsf{degree}(P, A, \phi) >\textsf{degree}(P, A, \psi)$                                             \\
    $\textsf{most}_\textsf{sup}(P, O, C, \phi)$     & $\mathcal{E} \slash \mathcal{A}$ & $\mathcal{P}, \mathcal{O}, \Phi \slash \mathcal{O}, \Phi \slash \mathcal{O}$ & $\lambda A. \textsf{degree}(P, A, \phi(O)) \geq \max_{x: C(x)} \textsf{degree}(P, A, \phi(x))$                    \\
    $\textsf{most}_\textsf{str}(P, \phi)$           & $\mathcal{E} \slash \mathcal{A}$ & $\mathcal{P}, \Phi$                                                          & $\lambda A. \textsf{degree}(P, A, \phi) \geq \alpha_{\textsf{most}} \cdot \theta_{P}$                                \\
    \bottomrule
    \end{tabular}
    \vspace{-3pt}
    \caption{Epistemic terms and definitions (\textsf{unlikely}, \textsf{less} and \textsf{least} are omitted due to space limits).}
    \vspace{3pt}
    \end{subtable}
    \begin{subtable}[b]{0.9\textwidth}
    \centering
    \begin{tabular}{@{}lllllllllll@{}}
    \toprule
    \multicolumn{10}{@{}l@{}}{\textbf{Thresholds} $\theta$}                                                                                                                                                                                                                                     & \textbf{Multipliers} $\alpha$  \\ \midrule
    $\textsf{believes}$ & $\textsf{certain}$ & $\textsf{uncertain}$ & $\textsf{likely}$ & $\textsf{unlikely}$ & $\textsf{could}$ & $\textsf{might}$ & $\textsf{may}$ & $\textsf{should}$ & $\textsf{must}$ & $\textsf{most}$ \\
    0.75                       & 0.95                      & 0.70                        & 0.70                     & 0.40                       & 0.20                    & 0.20                    & 0.30                  & 0.80                     & 0.95                   & 1.5                    \\ \bottomrule
    \end{tabular}
    \vspace{-3pt}
    \caption{Probability thresholds and multipliers (fitted against human data).}
    \vspace{3pt}
    \end{subtable}
    \vspace{-6pt}
    \caption{\textbf{Expressions in our epistemic language of thought (ELoT)}, including \textbf{(a)} epistemic terms and \textbf{(b)} probability thresholds $\Theta$. ELoT terms may have the following types: $\mathcal{E}$: epistemic formula, $\Phi$: base formula, $\Phi_\mathcal{F}$: function term, $\mathcal{P}$: predicate symbol, $\mathcal{A}$: agent, $\mathcal{O}$: object, $\mathcal{X \slash Y}$: function from $\mathcal{X} \to \mathcal{Y}$.}
    \label{tab:elot}
    \vspace{-6pt}
\end{table*}

\subsection{Interpreting Belief Sentences with an Epistemic Language of Thought}

To represent epistemic concepts in a way that mirrors the structure of natural language, we introduce a formal language (Table \ref{tab:elot}) as our epistemic language of thought.
Drawing upon the formal semantics of epistemic language \cite{hintakka1962knowledge,lassiter2010gradable}, we adopt a compositional degree-based semantics for epistemic concepts grounded in probability comparisons \cite{moss2015semantics,lassiter2017graded}.  This allows us to quantitatively evaluate an epistemic expression using probabilities inferred by our BToM module.

\subsubsection{Representing Epistemic Formulas}
\label{sec:sentence-representation}

We first define our non-epistemic base language --- a first order logic derived from the Planning Domain Definition Language \cite{mcdermott1998pddl}. Our language assumes a set of predicates $\mathcal{P}$ and functions $\mathcal{F}$ used to describe a set of objects $\mathcal{O}$. Predicates can be combined into formulas $\phi \in \Phi$ via logical operators or quantification. For example, \emph{``A key is in box 2''} can be represented as $\exists k. \textsf{key}(k) \land \textsf{inside}(k, \textsf{box2})$. Conceptually, a state $s$ of the environment (or an agent's mental representation of state $s$) is just a large formula: A conjunction of predicates which fully describe the state. We denote the truth value of $\phi$ in $s$ as $\llbracket \phi \rrbracket_s$.

On top of this base language $\Phi$, we introduce epistemic expressions $\varphi \in \mathcal{E}$ to model assertions of belief, knowledge, or modal qualifications (Table \ref{tab:elot}a). Following \citet{lassiter2017graded}, the semantics of these expressions are grounded in the probability function $\textbf{Pr}(A, \phi)$ --- the probability assigned by agent $A$ to formula $\phi$ --- and comparisons with term-specific thresholds (Table \ref{tab:elot}b). For example, the operator $\textsf{might}(\phi)$ takes a first-order formula $\phi\in\Phi$, returning a function $\lambda A . \textbf{Pr}(A, \phi) \geq \theta_{\textsf{might}}$. Combined with the operator $\textsf{believes}_\textsf{modal}(A, F) = F(A)$, we can express the claim that \emph{``$A$ believes it might be that $\phi$''} as $\varphi$ $=$ $\textsf{believes}_\textsf{modal}(A, \textsf{might}(\phi))$. This formula $\varphi$ can be lowered to the probability comparison $\textbf{Pr}(A, \phi) \geq \theta_\textsf{might}$, which uses $\theta_\textsf{might}$ as the threshold instead of $\theta_\textsf{believes}$ due to how our composition rules. We call this latter representation a \emph{lowered formula}.

A key aspect of ELoT expressions is that they represent epistemic concepts at a \emph{similar level of granularity} as our tested language (English), simplifying the mapping to ELoT propositions. In contrast, our lowered representation is just as expressive, but fails to match the structure of natural epistemic language. In our experiments, we show that translating into ELoT significantly improves the accuracy of semantic parsing.

\subsubsection{Translating Epistemic Language}
\label{sec:sentence-translation}

Natural language is varied and imprecise, with possibly many sentences $\sigma$ mapping to the same ELoT formula $\varphi$ (and vice versa). To handle this diversity, we perform semantic parsing via grammar-constrained LLM decoding \cite{shin2021constrained,scholak2021picard,willard2023efficient}, producing a flexible context-sensitive mapping between natural language and ELoT expressions \cite{wong2023word}. Specifically, we use sequential Monte Carlo (SMC)-based grammar-constrained sampling \citep{loula2025syntactic}, since it avoids the failure modes of beam search and greedy token-masking \cite{lew2023sequential}. We prompt an LLM (LLaMa 3.1 8B) with example translations $\mathcal{D}$ from English to ELoT and a sentence $\sigma$ to translate, using SMC to approximate the distribution over completions constrained to the ELoT grammar $G_\mathcal{E}$:
\vspace{-3pt}
\begin{align}
    \varphi &\sim P_\text{LLM}(\varphi | \sigma, \mathcal{D}, \varphi \in G_\mathcal{E})
\end{align}
We run SMC with $n_\sigma$ samples. This produces weighted samples $\{(\varphi_i, w_i)\}_{i=1}^{n_\sigma}$, and we use the top-weighted sample as our ELoT translation $\varphi$.\footnote{The full set of SMC samples can instead be used to handle ambiguous sentences with multiple plausible ELoT translations. We leave the study of this to future work.}

\subsection{Inferring and Evaluating Beliefs with a Bayesian Theory-of-Mind}
\label{sec:btom}

Our ELoT semantics reduces belief expressions to the probability $\textbf{Pr}(A, \phi)$ an agent $A$ assigns to a sentence $\phi$. But where does $\textbf{Pr}(A, \phi)$ come from? This is the function of our BToM module: By modeling the functional role that belief plays in guiding actions, along with the influence of perceptions on beliefs, an observer can infer what the agent thinks based on what the agent sees and does. Following the structure of Partially Observable Markov Decision Processes (POMDPs) \cite{kaelbling1998planning}, this theory of approximately rational agency can be formalized as a probabilistic generative model:
\vspace{-6pt}
\begin{alignat}{2}
\textit{Goal Prior:}& \quad g \sim P(g) \label{eq:goal-prior} \\
\textit{State Prior:}& \quad s_0 \sim P(s_0) \label{eq:state-prior} \\
\textit{Belief Prior:}& \quad b_0 \sim P(b_0 | s_0) \label{eq:belief-prior} \\
\textit{State Transition:} & \quad s_t \sim P(s_t | s_{t-1}, a_{t-1}) \label{eq:state-transition} \\
\textit{Belief Update:}& \quad b_t \sim P(b_t | s_t, b_{t-1}) \label{eq:belief-update} \\
\textit{Action Selection:}& \quad a_t \sim P(a_t | b_t, g) \label{eq:action-selection} \\
\textit{Observations:} & \quad o_t \sim P(o_t | s_t) \label{eq:observation-model}
\end{alignat}

\subsubsection{Modeling Perception and Action}
\label{sec:modeling-perception-and-action}

Two crucial aspects of our BToM module are how it models belief updating (Eq. \ref{eq:belief-update}) as the result of perception, and how it models goal-directed action given the agent's uncertain beliefs (Eq. \ref{eq:action-selection}). To model perception, we represent an agent's belief $b_t$ as a probability-weighted collection $\{(\tilde s_i, w_i)\}_{i=1}^{n_s}$ of possible environment states $\tilde s_i$ (which are represented in turn as collections of ELoT predicates). Given an observation of the environment $s_t$ (e.g. observing that a box is empty), the agent updates its belief by \emph{filtering} out inconsistent hypotheses $\tilde s_i$, setting $w_i = 0$.

As for goal-directed action, our model builds upon methods for epistemic planning \cite{bolander2017gentle} and belief-space planning in POMDPs \cite{littman1995learning}. Given a belief $b_t$, the agent engages in \emph{instrumental planning} to achieve their goal $g$, which requires achieving instrumental subgoals (e.g. picking up keys), but also gathering goal-relevant information (e.g. finding out if a key is in a certain box). We model this by assuming that the agent acts by approximately minimizing a \emph{cost-to-go estimate} $\hat Q_g(b_t, a)$: An estimate of the optimal cost $Q_g^*(b_t, a)$ of reaching $g$ after action $a$ starting from one's (uncertain) belief $b_t$. Action selection can thus be modeled by a Boltzmann distribution over these $\hat Q_g$ estimates:
\begin{equation}
    P(a_t | b_t, g) \propto \exp\left(-\beta \hat Q_g(b_t, a)\right)
    \label{eq:boltzmann-qmdp}
\end{equation}
To estimate $Q^*_g$ efficiently, we follow recent advances in inverse planning \cite{zhixuan2024pragmatic} by computing the $Q_\text{MDP}$ approximation \cite{hauskrecht2000value} of $Q_g^*$, averaging over the $Q$-values for each hypothesis $(\tilde s_i, w_i)$ in the belief $b_t$:
\begin{equation}
    \hat Q_g(b_t, a) = \textstyle\sum_{(\tilde s_i, w_i) \in b_t} w_i \cdot Q^*_g(\tilde s_i, a)
    \label{eq:qmdp}
\end{equation}
The cost-to-go $Q^*_g(s, a)$ from a known state $s$ can itself be efficiently estimated by searching for a shortest path to $g$ from $s$ \cite{monfort2015softstar}.

\subsubsection{Joint Inference of Goals and Beliefs}
\label{sec:joint-inference}

With this generative model, observers can jointly infer the agent's goal $g$, belief history $b_{0:T}$, and environment trajectory $s_{0:T}$ given observations of the agent's actions $a_{1:T}$ and partial observations $o_{0:T}$ of the environment:
\begin{align}
    &P(g, b_{0:T}, s_{0:T} | a_{1:T}, o_{0:T}) \propto     \label{eq:posterior} \\
    &\quad P(g, s_0, b_0) \textstyle\prod_{t=1}^T P(b_t, a_t, s_t, o_t | b_{t-1}, s_{t-1}) \nonumber
\end{align}

To ensure tractable posterior inference, we considered only the set of initial states $\mathcal{S}_0$ consistent with the initial observation $o_0$, and a discrete set $\mathcal{B}_0$ of possible beliefs $b_0$ sufficient to model comparative likelihood claims (e.g. \emph{``The key is more likely in box 1 than 2.''}). Specifically, we consider all beliefs formed by distributing $k$ particles across $n_s := |\mathcal{S}_0|$ states, resulting in $n_b := |\mathcal{B}_0| = \binom{n_s +  k - 1}{k}$  distributions. We then perform \emph{exact} Bayesian inference over all combinations of goals $g$, initial beliefs $b_0$, and states $s_0$, which we implement as a variant of Sequential Inverse Plan Search  \cite{zhi2020online} using the Gen probabilistic programming system \cite{cusumano2019gen}. More algorithmic details are provided in the Appendix.

\subsubsection{Evaluating Epistemic Sentences}
\label{sec:sentence-evaluation}

By inferring the agent's belief history $b_{0:T}$, we can compute the probability $\mathbf{Pr}(A, \phi)$ of a formula $\phi$ at time $t$ as the expected truth value under $b_t$:
\begin{equation}
    \mathbf{Pr}(A, \phi) = \textstyle\sum_{(\tilde s_i, w_i) \in b_t} w_i \cdot \llbracket \phi \rrbracket_s
    \label{eq:prob_of_formula}
\end{equation}
We can thus evaluate a epistemic formula $\varphi$ given a belief $b_t$ and environment state $s_t$ by replacing $\mathbf{Pr}$ terms with their values, then determining the truth of the resulting expression in state $s$. We denote this operation by $\llbracket \varphi \rrbracket_{(s_t, b_t)}$.

However, observers do not have access to the true $s_t$ or $b_t$, only inferences about them. As such, we model human evaluation of a sentence $\varphi$ as a probabilistic judgment given their inferences:
\begin{multline*}
    P(\llbracket \varphi \rrbracket_{(s_t, b_t)}| a_{1:T}, o_{0:T}) \\
    = \mathbb{E}_{s_t, b_t \sim P(s_t, b_t | a_{1:T}, o_{0:T})}\left[\llbracket \varphi \rrbracket_{(s_t, b_t)}\right]
\end{multline*}
While these judgements are made after observing actions up to time $T$, $\varphi$ may be \emph{retrospectively} evaluated as a description of the agent's beliefs at any $t \in [0, T]$, with $t$=$0$ and $t$=$T$ corresponding to initial and current beliefs respectively.

Following \citet{ying2024grounding}, we also assume that people provide ratings as if they have a uniform prior $U_\varphi$ about the truth of $\varphi$. Under this prior, ratings of $\varphi$ can be interpreted as a \emph{normalized likelihood} $\bar L(\llbracket \varphi \rrbracket_{(s_t, b_t)} | a_{1:T}, o_{0:T})$, which measures the likelihood of statement $\varphi$ relative to its negation $\neg\varphi$. In other words, we assume that humans rate an epistemic claim more highly when they have \emph{evidence} for it. With no evidence, $\bar L(\llbracket \varphi \rrbracket_{(s_t, b_t)}| a_{1:T}, o_{0:T}) = 0.5$. Results investigating the importance of this assumption can be found in the Appendix.

\section{Experiments}

To evaluate our model on a diverse dataset of epistemic language (Table \ref{tab:dataset_overview}), we conducted a two-part human experiment. We first recruited participants to write English sentences describing the current and initial beliefs of a player character as it navigated a gridworld puzzle that required finding keys hidden in boxes. Next, we asked two groups of participants to rate how likely these sentences were to be true given the player's behavior, with one group rating sentences about the player's current beliefs, and the other rating sentences about initial beliefs.

With this data, we evaluated our model by: (i) assessing the \textbf{translation accuracy of our ELoT module}, investigating the importance of our ELoT representation (vs. the lowered form) and its impact on grammar-constrained LLM decoding; (ii) testing our \textbf{full LaBToM model} in its ability to capture human interpretation of epistemic language, comparing human ratings against LaBToM inferences. Model and experiment code is available at \url{https://osf.io/xq9c4/}.

\subsection{Scenario Construction}

We constructed 20 scenarios in the Doors, Keys, \& Gems environment with varied maze designs and item locations (Figure \ref{fig:qualitative}). In each scenario, there were 4 goal gems with different shapes (triangle, square, hexagon, circle), some of which were locked behind doors. Scenarios also had 2 to 3 boxes with up to 2 colored keys among them. The player's actions were varied across scenarios to elicit inferences about a diversity of epistemic states, such as ignorance about key locations or false confidence about the location of a key.

\begin{table*}[t]
    \centering
    \footnotesize
    \begin{tabular}{@{}llll@{}}
    \toprule
    \textbf{Factor}  & \textbf{Description}                                                                                        & \multicolumn{2}{l}{\textbf{Count}}  \\
                     & \textit{Examples}                                                                                           & \textit{Current} & \textit{Initial} \\ \midrule
    Possibility      & Sentences with modal verbs such as might, could and must.                                                   & 66 / 241         & 46 / 228         \\[-1pt]
                     & {\scriptsize \emph{The player believes box 1 may contain a blue key or a red key.}}                         &                  &                  \\[-1pt]
                     & {\scriptsize \emph{The player believes if the red key is not in box 2 then it must be in box 3.}}           &                  &                  \\[1pt]
    Probability      & Sentences with probability expressions such as likely, uncertain, etc.                                      & 28 / 241         & 28 / 228         \\[-1pt]
                     & {\scriptsize \emph{The player thought that box 1 was most likely to contain a red key.}}                    &                  &                  \\[-1pt]
                     & {\scriptsize \emph{The player is unsure what color the key in box 2 will be.}}                              &                  &                  \\[1pt]
    Compositionality & Sentences that embed compound propositions (conjunctions, disjunctions, etc.).                              & 49 / 241         & 69 / 228         \\[-1pt]
                     & {\scriptsize \emph{The player thinks that there's more likely to be a red key in box 1 or 3 than box 2.}}   &                  &                  \\[-1pt]
                     & {\scriptsize \emph{The player believes that if box 1 does not have a blue key, then box 3 has a blue key.}} &                  &                  \\[1pt]
    Knowledge        & Sentences that make knowledge or ignorance claims.                                                          & 17 / 241         & 20 / 228         \\[-1pt]
                     & {\scriptsize \emph{The player already knows for sure there is no key in box 1 or box 2.}}                   &                  &                  \\[-1pt]
                     & {\scriptsize \emph{The player did not know if box 2 contained a red key.}}                                  &                  &                  \\ \bottomrule
    \end{tabular}
    \vspace{-6pt}
    \caption{Overview of our dataset of 464 human-written epistemic sentences, broken down by factors.}
    \vspace{-9pt}
    \label{tab:dataset_overview}
\end{table*}

\subsection{Collecting Epistemic Language}

In the first part of the experiment, we recruited 42 US participants via Prolific (mean age: 36.02, SD: 10.1; 16 women, 26 men). Following a tutorial, participants watched 10 scenario animations, with each stopping before the player reached their goal and all relevant keys were revealed. Participants were then asked to write at least two sentences about the player's likely beliefs at the end of the scenario (\emph{current beliefs}), and another two sentences about the player's beliefs at the start of the scenario (\emph{initial beliefs}). 
To ensure that these sentences focused on beliefs about the environment, we instructed participants to describe the player's beliefs about the contents of the boxes.  We excluded 8 participants for failing to follow these instructions, and about one third of remaining sentences (see Appendix).
This process left us with 241 (228) statements about current (initial) beliefs, which were then annotated with factors by two experimenters. Table \ref{tab:dataset_overview} illustrates the diversity of language we collected.

\subsection{Evaluating Epistemic Language}

For the next part of our experiment, we recruited 94 US participants via Prolific to evaluate current belief statements (mean age = 35.7, SD = 12.4, 67 women, 27 men), and another 104 US participants to evaluate initial belief statements (mean age = 35.27, SD = 11.7, 69 women, 33 men, 2 non-binary). Each participant was shown 10 out of 20 scenario animations, and was asked to rate the goals and beliefs of the player at several judgment points during each animation. For goals, participants were shown a checkbox for each gem, and asked to select all gems likely to be the agent's goal. This served as both an attention check and an additional data source for model validation. For beliefs, participants were shown 2 belief statements selected from our dataset of human-written statements, and asked to rate how likely each statement was on a scale from 1 to 7. These ratings were normalized between 0 and 1 for our analysis. We excluded 7 participants from the current belief condition and 5 from the initial belief condition for low outlying scores on the goal inference subtask.

\subsubsection{Statement Selection}

To ensure that the belief statements evaluated by our participants were (i) diverse and (ii) rated enough times to ensure sufficient statistical power (88\% power at Cohen's $d=0.8$), we selected 5 statements per scenario (3 plausible, 2 implausible) from our full dataset of statements. The plausible statements were chosen by sampling many sets of 3 statements out of all those written for a scenario, then selecting the set that scored highest on a diversity metric derived from the factors in Table \ref{tab:dataset_overview} (see Appendix for details). We then manually added 2 more statements that were originally written for other scenarios, and which we evaluated to be implausible descriptions of the target scenario. Participants were shown 2 of these 5 statements at random in each scenario.

\subsection{ELoT Parser Evaluation}

We evaluated the suitability of ELoT formulas as representations of epistemic sentences by investigating (i) the impact of using ELoT as the translation target (instead of the lowered form, which lacks epistemic operators besides $\mathbf{Pr}$) and (ii) the value of the associated ELoT grammar $G_\mathcal{E}$ in constraining LLM outputs.

To do so, we created gold translations of the subset of statements selected for human evaluation (125 statements) into ELoT formulas and their lowered forms. We then compared the translations produced by grammar-constrained SMC decoding ($n_\sigma$$=$$10$ samples) of LLaMa 3.1 8B against these gold translations in terms of both strict equality and approximate semantic equivalence (i.e. cases of reasonable alternative translations of the original English statement). As baselines, we used unconstrained self-consistency sampling \cite{wang2023selfconsistency} with LLaMa 3.1 8B and the instruction-tuned Gemini Flash 8B, sampling $n_\sigma$$=$$10$ times and taking the majority answer. All methods used the same prompt format (see Appendix), with 36 example translations.

\subsection{LaBToM Fitting and Evaluation}
\label{sec:model-evaluation}

We evaluated our LaBToM model on all 20 scenarios, producing normalized likelihood scores for the 5 current and 5 initial belief statements per scenario. We then computed Pearson's $r$ between these scores and average human ratings. We fit our model parameters to maximize $r$, fitting the belief thresholds $\Theta := (\theta_\textsf{believes}, \theta_\textsf{could}...)$ via coordinate ascent, and the inverse temperature $\beta$ via grid search (see Appendix for robustness analyses). This produced the fitted values for $\Theta$ in Table \ref{tab:elot}, and $\beta = 2^{3/2}$. For the set of possible initial agent beliefs $\mathcal{B}_0$, we fixed the number of belief particles to $k = 3$ to ensure tractable exact inference.

Alongside this direct comparison with human-provided ratings, we evaluated our model on the full dataset of 469 human-written statements by pairing each statement with either the scenario it was written for (\emph{in-context}) or 1--2 other scenarios with the same map layout but distinct agent trajectories (\emph{out-of-context}). This allowed us to compare each statement's in-context normalized likelihood $\bar{L}$ with its (average) out-of-context likelihood score. Reasonable models of epistemic language interpretation should assign higher likelihood scores to most statements when they are evaluated in-context vs. out-of-context.

\subsection{Baselines}

To assess the import of a sufficiently rich theory of mind for epistemic language understanding, we evaluated several ablations of our model under simplified assumptions about the agent's beliefs or planning abilities.
We also evaluated two state-of-the-art multi-modal LLMs, thereby testing the degree to which grounded evaluation of epistemic language can be achieved with sufficient scale:

\paragraph{True Belief.} The True Belief ablation assumes that the observed agent has fully accurate beliefs about the environment (i.e. they already know where all the keys are located), equivalent to the full model from \citet{ying2024grounding}. The observer starts with a uniform prior over these true beliefs.

\paragraph{Non-Planning.} The Non-Planning ablation as-sumes that the agent is incapable of planning towards instrumental subgoals (such as keys), and instead optimizes the heuristic of moving physically closer to the goal. This is implemented by using the Manhattan distance to the goal as the agent's cost-to-go estimate $\hat Q_g$.

\paragraph{Multi-modal LLMs.} We use GPT-4o (text \& image input, \texttt{gpt-4o-2024-05-13}) and Gemini 1.5 Pro (text \& image or video input) as multi-modal LLM baselines, providing them the same instructions as human participants (see Appendix for a unimodal baseline). Each baseline was run 3 times with a temperature of 1.0. In addition to providing an image or video of the scenario showing the actions up to each judgment point, we used the following prompting methods:

\textbf{\emph{Plans:}} The prompt plainly describes the agent's actions over time (e.g. \emph{the player moves right five times}), and also describes the agent's observations (e.g. \emph{the player opens box 1 and finds a red key}).

\textbf{\emph{Narratives:}} A rich narrative of agent behavior is included in the prompt, providing key information about the scene (e.g. which gems are locked behind doors, which keys are visible), while also describing the agent's movements in relation to relevant objects (e.g. \emph{the player moves right five times, going past box 1 towards box 2}).

\textbf{\emph{Few-Shot Prompting:}} After describing the plan or narrative, we provide the LLM with human ratings for the 4 other statements tied that scenario, before querying its rating for the target statement.

\section{Results}

\begin{table}[t]
    \centering
    \footnotesize
\begin{tabular}{lllll}
\hline
\multicolumn{1}{c}{\textbf{Parser}} & \multicolumn{4}{c}{\textbf{Translation Accuracy} (s.e.)}                                                                                                                                                                         \\
                                    & \multicolumn{2}{c}{\textit{\textbf{ELoT}}}                                                                  & \multicolumn{2}{c}{\textit{\textbf{Lowered}}}                                                               \\
                                    & \multicolumn{1}{c}{\textit{Exact}}                   & \multicolumn{1}{c}{\textit{Equiv.}}                  & \multicolumn{1}{c}{\textit{Exact}}                   & \multicolumn{1}{c}{\textit{Equiv.}}                  \\ \hline
\multicolumn{5}{l}{\textbf{Grammar-Constrained SMC} \cite{loula2025syntactic}}                                                                                                                                                                                                            \\
LLaMa 3.1 8B                        & {\renewcommand{\arraystretch}{0.85}\begin{tabular}[c]{@{}l@{}}\textbf{0.81}\\ {\scriptsize(.03)}\end{tabular}} & {\renewcommand{\arraystretch}{0.85}\begin{tabular}[c]{@{}l@{}}\textbf{0.91}\\ {\scriptsize(.03)}\end{tabular}} & {\renewcommand{\arraystretch}{0.85}\begin{tabular}[c]{@{}l@{}}0.34\\ {\scriptsize(.04)}\end{tabular}} & {\renewcommand{\arraystretch}{0.85}\begin{tabular}[c]{@{}l@{}}0.42\\ {\scriptsize(.04)}\end{tabular}} \\ \hline
\multicolumn{5}{l}{\textbf{Unconstrained Sampling + Majority Vote}}                                                                                                                                                                                                             \\
LLaMa 3.1 8B                        & {\renewcommand{\arraystretch}{0.85}\begin{tabular}[c]{@{}l@{}}0.65\\ {\scriptsize(.04)}\end{tabular}} & {\renewcommand{\arraystretch}{0.85}\begin{tabular}[c]{@{}l@{}}0.69\\ {\scriptsize(.04)}\end{tabular}} & {\renewcommand{\arraystretch}{0.85}\begin{tabular}[c]{@{}l@{}}0.32\\ {\scriptsize(.04)}\end{tabular}} & {\renewcommand{\arraystretch}{0.85}\begin{tabular}[c]{@{}l@{}}0.36\\ {\scriptsize(.04)}\end{tabular}} \\ 
Gemini Flash 8B                     & {\renewcommand{\arraystretch}{0.85}\begin{tabular}[c]{@{}l@{}}0.69\\ {\scriptsize(.04)}\end{tabular}} & {\renewcommand{\arraystretch}{0.85}\begin{tabular}[c]{@{}l@{}}0.80\\ {\scriptsize(.03)}\end{tabular}} & {\renewcommand{\arraystretch}{0.85}\begin{tabular}[c]{@{}l@{}}0.60\\ {\scriptsize(.04)}\end{tabular}} & {\renewcommand{\arraystretch}{0.85}\begin{tabular}[c]{@{}l@{}}0.68\\ {\scriptsize(.04)}\end{tabular}} \\ \hline
\end{tabular}
    \caption{\textbf{Translation accuracy for ELoT vs. lowered formulas as the translation target}, compared across sampling methods and LLMs in terms of exact equality (\textit{Exact}) or semantic equivalence (\textit{Equiv.}). ELoT serves as a better target \emph{and} constraint for translation.}
    \label{tab:parser}
    \vspace{-9pt}
\end{table}

\begin{table*}[t]
\centering
\footnotesize
\begin{tabular}{@{}lll@{}}
\toprule
 & \textbf{Parser} & \textbf{Example}                                                                                         \\ \midrule
\textbf{English} & & The player believes that box 3 is empty.                                                                \\
\textbf{ELoT} & \textit{Gold} & {\scriptsize \textsf{believes(player, formula(empty(box3)))}}                                                                \\
& \textit{Grammar-Cons.} & {\scriptsize \textsf{believes(player, formula(empty(box3)))}}                                                              \\
& \textit{Unconstrained} & {\scriptsize \textsf{believes(player, {\color{red}empty(box3)})}} {\color{red} \textbf{(Syntax Error)}}                                                               \\
\textbf{Lowered} & \textit{Gold} & {\scriptsize \textsf{>=(prob\_of(player, empty(box3)), threshold(believes))}}                                                                \\
& \textit{Grammar-Cons.} & {\scriptsize \textsf{{\color{magenta}and(}>=(prob\_of(player, empty(box3)), threshold(believes)), {\color{magenta} empty(box3))}}} {\color{magenta} \textbf{(Semantic Error)}}                                                                \\
& \textit{Unconstrained} & {\scriptsize \textsf{>=(prob\_of(player, empty(box3)), {\color{red} believes})}} {\color{red} \textbf{(Syntax Error)}}                                                                \\ \bottomrule
\end{tabular}
\vspace{-6pt}
\caption{Example translations from English to ELoT and lowered formulas using LLaMa 3.1 8B.}
\vspace{-6pt}
\label{tab:translation_examples}
\end{table*}

\paragraph{ELoT is a superior translation target and guide for (grammar-constrained) LLM parsing.} We report translation accuracy in Table \ref{tab:parser}. In confirmation of the idea that our ELoT formalism captures the semantics of belief sentences at the right level of granularity, we find that translating English into ELoT is up to 2.4 times as accurate as translating into the lowered form, even though they have the same expressive capacity. ELoT also increases the benefit of grammar-constrained SMC decoding, leading to an improvement of 0.20 in equivalence accuracy for LLaMa 3.1 8B. In contrast, using the lowered grammar leads to only an 0.06 improvement over unconstrained generation. We show example translations and errors for each representation in Table \ref{tab:translation_examples}. These findings highlight the importance of ``language-of-thought'' representations that are aligned with the structure of natural language, in line with work tying language use to the structure and acquisition of compositional concepts \cite{wong2022identifying}.

\begin{table}[t]
\centering
\footnotesize
\begin{tabular}{llll}
\hline
\multicolumn{1}{c}{\textbf{Model}} & \multicolumn{3}{c}{\textbf{Human Correlation $r$} (s.e.)}                 \\
                                   & \textit{All}         & \textit{Current}     & \textit{Initial}     \\ \hline
\textbf{LaBToM}                    &                      &                      &                      \\
Full (ours)                        & \textbf{0.76 (0.01)} & \textbf{0.78 (0.01)} & \textbf{0.72 (0.01)} \\
Non-Planning                       & 0.40 (0.01)          & 0.58 (0.01)          & 0.07 (0.01)          \\
True Belief                        & 0.10 (0.01)          & 0.09 (0.01)          & 0.09 (0.02)          \\ \hline
\multicolumn{4}{l}{\textbf{GPT-4o}  \cite{openai2023gpt4}}           \\
I+Na+FS                            & 0.52 (0.01)          & 0.59 (0.01)          & 0.41 (0.01)          \\
I+Na                               & 0.48 (0.01)          & 0.52 (0.01)          & 0.41 (0.01)          \\
I+Pl                               & 0.28 (0.01)          & 0.32 (0.01)          & 0.18 (0.01)          \\ \hline
\multicolumn{4}{l}{\textbf{Gemini 1.5 Pro} \cite{geminiteam2024gemini}}               \\
V+Na+FS                            & 0.23 (0.01)          & 0.28 (0.01)          & 0.14 (0.02)          \\
I+Na+FS                            & 0.22 (0.01)          & 0.29 (0.01)          & 0.11 (0.02)          \\ \hline
\multicolumn{4}{l}{{\scriptsize I - Image, V - Video, Pl - Plans, Na - Narratives, FS - Few-Shot}}     
\end{tabular}
\vspace{-6pt}
\caption{\textbf{Model correlations with human ratings of epistemic language}. LaBToM correlates strongly with humans, whereas multimodal LLMs struggle to do so.}
\vspace{-12pt}
\label{tab:performance_summary}
\end{table}

Overall, grammar-constrained SMC decoding into ELoT with LLaMa 3.1 8B performs the best, achieving 91\% equivalence with our gold translations. Upon inspection (see Appendix), we find that unconstrained sampling often leads to syntax errors not corrected for by majority voting. Gemini Flash 8B also exhibits low sample diversity typical of instruction tuning, such that majority voting gives no improvement. In comparison, grammar-constrained SMC produces not just accurate translations, but a \emph{distribution} over reasonable translations, since SMC performs posterior sampling rather than optimization \cite{loula2025syntactic,lew2023sequential}. In future work, this could be used to more precisely model the interpretation of ambiguous epistemic language.

\begin{figure}[t]
    \centering
    \includegraphics[width=.98\columnwidth]{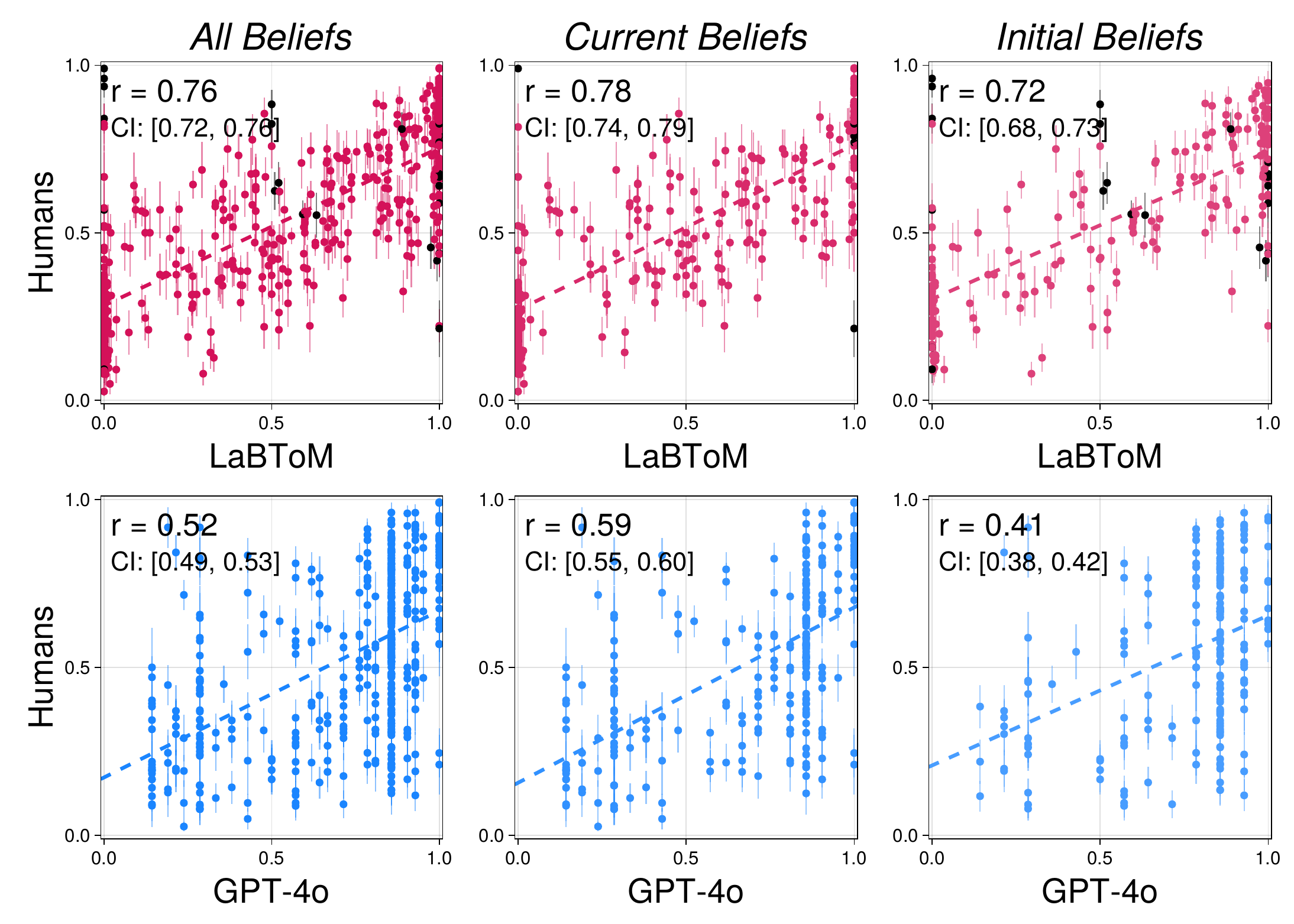}
    \vspace{-2pt}
    \caption{\textbf{Human correlation plots for LaBToM and GPT-4o.} Black dots are from statements not accurately translated to ELoT. LaBToM provides a much stronger qualitative fit with human ratings compared to the best GPT-4o baseline, which fails to use the full scale.}
    \label{fig:correlation}
    \vspace{-9pt}
\end{figure}

\paragraph{LaBToM correlates highly with human ratings of epistemic statements.} As Table \ref{tab:performance_summary} shows, our full model produces statement scores that correlate highly with human ratings ($r$ = $0.76$) across both current and initial belief conditions. In the Appendix, we show results using the gold ELoT translations ($r$ = $0.81$), and a per-factor breakdown. Plotting average human judgments against LaBToM inferences (Figure \ref{fig:correlation}, Row 1), we also find a strong qualitative fit: Factoring out incorrect ELoT translations (black dots), LaBToM generally assigns high or low ratings to sentences when humans do, using the ends of the scale as appropriate. In contrast, our ablated models do poorly, either because they fail to track how the player updates their beliefs (True Belief, Current $r$ = $0.09$), or fail to infer the player's initial beliefs (Non-Planning, Initial $r$ = $0.07$). These findings hold even when the parameters of our full model are adversarially optimized (see Appendix).

\begin{figure*}[th!]
    \centering
    \includegraphics[width=\textwidth]{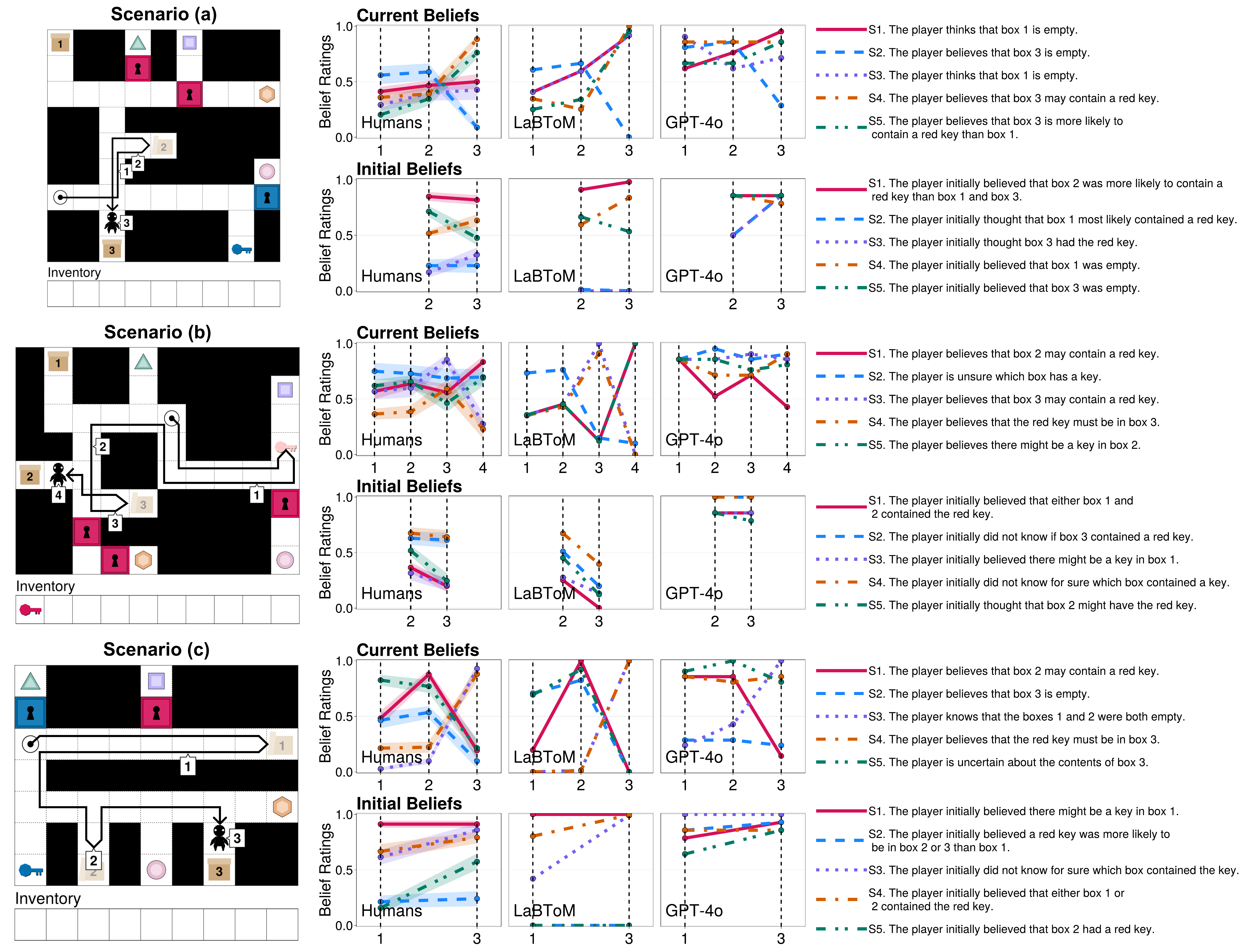}
    \caption{\textbf{Step-by-step ratings by humans and models across three scenarios}. Judgement points are annotated on each map, and show the player's location \emph{before} opening the nearest box. Keys picked up along the way are shown in light colors. Our model largely matches human responses qualitatively and quantitatively, unlike GPT-4o.}
    \label{fig:qualitative}
    \vspace{-6pt}
\end{figure*}

\paragraph{SotA multimodal LLMs struggle at grounded evaluation of epistemic language.} The multi-modal LLM baselines also perform less well than LaBToM despite extensive prompting, with the strongest LLM baseline (GPT-4o with images, narratives, and few-shot prompting) achieving human correlations of only 0.59 and 0.41 respectively. Increasing information in the prompt improves performance, although the Gemini models do poorly regardless, even with full scenario videos. We find that LLM performance is lower on initial belief sentences, suggesting that they are better at tracking how agents' current beliefs change, but worse at inferring past beliefs.

Figure \ref{fig:correlation} illustrates the difference between our model and the best performing GPT-4o baseline (I+Na+FS) in greater detail. Despite few-shot prompting, GPT-4o tends to assign ratings of 0.8 or more to many statements that humans find quite unlikely, suggesting that LLMs struggle to account for evidence \emph{against} a belief claim.

\paragraph{LaBToM captures how human evaluations of epistemic language change with agent behavior.} Our model predicts that human evaluations of epistemic sentences should change systematically as they gain more information about an agent's percepts and actions. As Figure \ref{fig:qualitative} illustrates, this is what we find. When the player sees that a box is empty, both humans and our model sharply decrease their ratings for statements claiming that the player believes a key is in that box (e.g. Fig. \ref{fig:qualitative}b, S4 Current). When the player approaches one box instead of another, both humans and LaBToM gain confidence in statements about the \emph{relative likelihood} of key locations (Fig. \ref{fig:qualitative}a, S5 Current). 

We analyze just one scenario (Figure \ref{fig:qualitative}a) in detail. Here, the player first walks upwards away from box 3 (Judgment Point 1), leading both humans and LaBToM to assign scores of greater than 0.5 to \emph{``The player believes that box 3 is empty''} and \emph{``The player initially believed that box 3 was empty''}. In contrast, the modal sentence \emph{``The player believes that box 3 may contain a red key''} is rated lower. This is because if the player did believe that box 3 might have the (red) key, then it is more likely that they would have looked in box 3. However, this does not occur.

After the player opens box 2 and finds it empty, then walks back down towards box 3 (Judgment Point 3), both humans and our model decrease their confidence in the statement \emph{``The player believes that box 3 is empty''}. They also decrease confidence in \emph{``The player initially believed that box 3 was empty''}, but \emph{less sharply}. This is because there are at least two possibilities consistent with the player's actions: They could have firmly believed that box 3 was empty, or they could have just believed that box 3 was \emph{less likely} to contain the relevant key than box 2, without all-out believing that it was empty. This ability to understand beliefs about relative likelihood is made explicit by how ratings change over time for \emph{``The player believes that box 3 is more likely to contain a red key than box 1.''} Consistent with the principle of rational action, both humans and LaBToM assign a high score to this sentence once the player goes back to box 3, forgoing box 1.

Across a range of other settings (Figure \ref{fig:qualitative}b--c), our model largely captures fine-grained changes in how people evaluate a variety of epistemic expressions, including modal sentences, ignorance claims, and expressions of uncertainty. Unlike prior BToM models that lack language-like belief representations \cite{baker2017rational}, LaBToM also distinguishes observer and agent uncertainty, assigning high ratings to claims that the agent is uncertain (Figure \ref{fig:qualitative}c, S5 Cur.), and vice versa. In contrast, the best LLM baseline (GPT-4o, I+Na+FS) often fails to adjust its ratings in the same direction as humans, while rating implausible statements highly. We discuss these results more in the Appendix, alongside cases where our model comes apart from humans. 

\begin{table}[t]
\centering
\footnotesize
\begin{tabular}{@{}lll@{}}
\toprule
\multicolumn{1}{c}{\textbf{Model}} & \multicolumn{1}{c}{\textbf{Statement Likelihoods}} & \textbf{Accuracy}     \\
                                   & \textit{In-Ctx / Out-of-Ctx / Diff.}               & \textit{In vs. Out}  \\ \midrule
\multicolumn{3}{c}{\textit{\textbf{Current Beliefs} (241 statements)}}                                                          \\ \midrule
LaBToM (ours)                      & \textbf{0.78 / 0.48 / +0.30 (0.03)}                & \textbf{0.71 (0.03)} \\
Non-Planning                       & 0.70 / 0.54 / +0.16 (0.02)                         & 0.65 (0.03)          \\
True Belief                        & 0.35 / 0.35 / +0.01 (0.01)                         & 0.11 (0.02)          \\
GPT-4o (I+Na)                     & 0.80 / 0.62 / +0.18 (0.03)                         & 0.59 (0.03)          \\ \midrule
\multicolumn{3}{c}{\textit{\textbf{Initial Beliefs} (228 statements)}}                                                          \\ \midrule
LaBToM (ours)                      & \textbf{0.73 / 0.51 / +0.22 (0.03)}                & \textbf{0.70 (0.03)} \\
Non-Planning                       & 0.70 / 0.70 / -0.01 (0.01)                         & 0.39 (0.03)          \\
True Belief                        & 0.51 / 0.50 / +0.00 (0.01)                         & 0.16 (0.02)          \\
GPT-4o (I+Na)                     & 0.76 / 0.68 / +0.07 (0.02)                         & 0.38 (0.03)          \\ \bottomrule
\end{tabular}
\caption{\textbf{In vs. out-of-context statement evaluation.} LaBToM most accurately distinguishes when epistemic language is evaluated in vs. out-of-context, assigning significantly higher scores in-context (s.e. in brackets).}
\label{tab:likelihood_comparison}
\vspace{-6pt}
\end{table}

\paragraph{LaBToM distinguishes in-context and out-of-context epistemic language.} To investigate how our model generalizes to a larger set of epistemic expressions, we performed the in-context vs. out-of-context likelihood comparison described in Section \ref{sec:model-evaluation} for our full dataset of 469 sentences. We tested the full LaBToM model, ablations and the best applicable LLM baseline from Table \ref{tab:performance_summary} (GPT-4o, I+Na). Results are shown in Table \ref{tab:likelihood_comparison}. For both current and initial belief statements, we find that LaBToM assigns significantly higher scores when a statement is evaluated in-context vs. out-of-context, correctly classifying the context about 70\% of the time by assigning a strictly higher in-context score. On closer inspection, many statements that are incorrectly classified turn out to be plausible in either context, resulting in equal or close-to-equal scores. The ablated models and GPT-4o perform significantly worse than LaBToM, especially for initial beliefs.

\section{Discussion}

Our experiments show that, similar to humans, our LaBToM model is able to coherently interpret and adjust its evaluations of natural language statements about agents' beliefs, whereas state-of-the-art multimodal LLMs struggle with this task. This ability is mediated by our ELoT representation and semantic parser: Without a compositional representation of epistemic concepts that is aligned with the structure of natural epistemic language, we find that translation accuracy drops significantly, even when using a grammar-constrained LLM parser. We also find that LaBToM largely distinguishes in vs. out-of-context sentence usage on a large and diverse set of crowd-sourced epistemic language, indicating the generalizability of our approach.

That said, our model is not without limitations. As we discuss at greater length in the Appendix, LaBToM's outputs depart from human judgments in several interesting ways, suggesting the need to account for (i) contextual adaptation of probability thresholds via pragmatic reasoning \cite{rudin2016deriving,schuster2020know}, (ii) the role of justification in human's intuitive evaluation of knowledge claims \cite{alston1989epistemic}, and (iii) bounded human reasoning about logical implications \cite{smets2018effort}.

LaBToM is also an ideal observer model that does not scale readily to large belief spaces, leaving open how humans \emph{tractably} infer and evaluate claims about others' beliefs \cite{van2008tractable}, perhaps by focusing on \emph{occurent} beliefs \cite{bartlett2018occurrent} that are relevant to others' goals. Finally, LaBToM is a model of how people \emph{interpret} epistemic language, but full understanding also includes the ability to \emph{produce} such language. This could potentially be achieved by inverting the ELoT module of our model, using it to translate salient or conversationally-relevant inferences about an agent's beliefs into natural language. By extending our model in this way, we stand to gain an even richer account of what it means to understand epistemic language.

\section*{Acknowledgments}

We thank our colleagues Cedegao Zhang, Brian Leahy, and Megan Wei for helpful discussions in the development of this project, and Ben LeBrun for his help in applying grammar-constrained SMC decoding to our setting. We also thank Jacob Andreas, Cedegao Zhang, Liam Bright, Daniel Lassiter, our reviewers, and our action editor for their valuable feedback on this paper.

This work was funded in part by the DARPA Machine Common Sense, AFOSR, and ONR Science of AI programs, along with the MIT-IBM Watson AI Lab, the Siegel Family Foundation, and an anonymous donor. Tan Zhi-Xuan is supported by an Open Philanthropy AI Fellowship.

\bibliography{tacl2021}
\bibliographystyle{acl_natbib}

\appendix
\onecolumn

\renewcommand\thefigure{\thesection\arabic{figure}}    
\setcounter{figure}{0}

\renewcommand\thetable{\thesection\arabic{table}}    
\setcounter{table}{0}

\section{Dataset Collection}

\subsection{Experimental Procedure}

\begin{figure*}[h!]
    \centering
    \includegraphics[width = 0.75\textwidth]{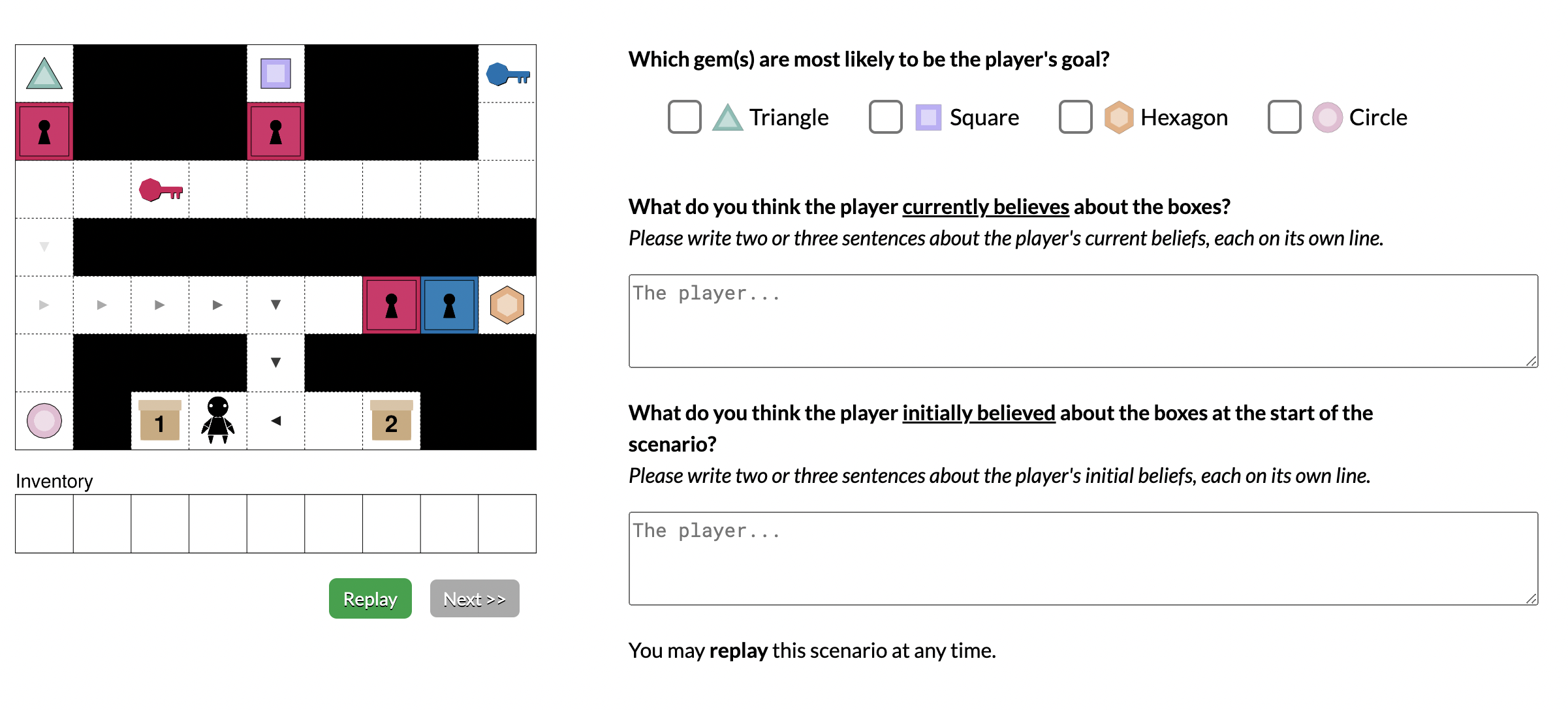}
    \includegraphics[width = 0.75\textwidth]{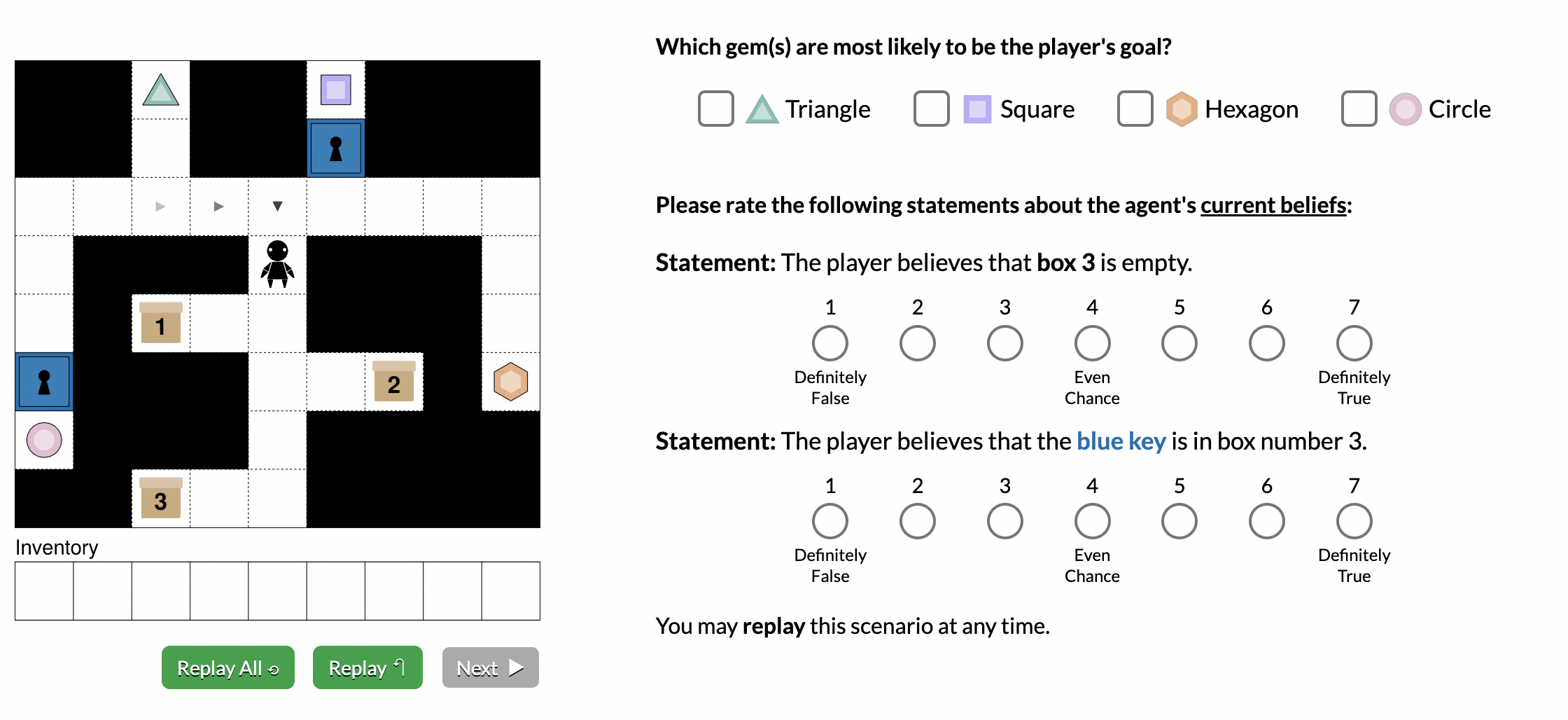}
    \caption{Interfaces used for collecting (top) and evaluating (bottom) epistemic language.}
    \label{fig:interface}
\end{figure*}

The interface used for collecting the statements and evaluating statements are shown in Figure \ref{fig:interface}. Participants first completed a tutorial that explained the task and experimental interface, then answered 5 comprehension questions before proceeding to the main experiment. In the main experiment, they were shown 10 out of the 20 stimuli in a randomized order. 

To incentivize accurate but calibrated responses, participants were rewarded for accurately guessing the true goal. Specifically, they earned $1/N$ bonus points if they selected $N$ goals out of which one was the true goal, but $0$ points if none of their selected goals was the true goal. Participants were paid US\$1 for every 40 bonus points they earned, on top of a base pay of US\$15/hr.

\subsection{Statement Post-Processing and Annotation}

Once statements were collected, two experimenters independently annotated whether each statement was valid for inclusion. We excluded invalid sentences based on two criteria: (i) whether the statement had the right tense (present for current beliefs, past for initial beliefs) and (ii) whether the statement referred to beliefs about the boxes. We also corrected minor grammatical errors and normalized statements to the form \emph{``The player + [believes/knows/thinks/expects/is sure/is uncertain, etc.]...``} for current beliefs and \emph{``The player initially + [believed/knew/thought/expected/was sure/uncertain, etc.]...''} for initial beliefs.

After filtering and normalization, two experimenters independently annotated the statements based on four factors: possibility, probability, compositionality, and knowledge. These factors are not mutually exclusive, so a statement could be annotated with any combination of factors. The codes for possibility were [\textit{may, might, can, could, should, must, none}] and the codes for probability were [\textit{certain, uncertain, likely, unlikely, none}]. The codes for compositionality and knowledge were binary [0, 1]. The annotators agreed on 95\% of the codes and discussed to resolve their differences.

\subsection{Selecting Diverse Statements for Human Evaluation}
After annotation, we selected a set of 3 plausible and 2 implausible statements per scenario for evaluation by human raters. Implausible statements were manually selected from other scenarios to be implausible in their scenario of evaluation. To select the 3 plausible statements, we sampled 100 subsets of 3 statements out of all statements written for that scenario, then computed a diversity score for each set $S$:
{\small
\begin{align}
    \text{Score}(S) = \frac{1}{4} |S_\text{possibility}| + \frac{1}{4} |S_\text{probability}| + \frac{1}{4} |S_\text{compositionality}| + \frac{1}{4} |S_\text{knowledge}|
\end{align}
}where $|S_\text{possibility}| $ indicates the number of unique possibility codes in set $S$ and vice versa. We then chose the set with the highest diversity score among the 100 sampled sets.

\section{Model Configuration}

\subsection{Belief-Space Sequential Inverse Plan Search (BSIPS)}

Our BToM inference algorithm is a belief-space policy variant of Sequential Inverse Plan Search (SIPS, \citet{zhi2020online}), which uses policies to evaluate action likelihoods as in recent extensions of SIPS \cite{zhixuan2024pragmatic,zhixuan2024infinite}. Perhaps surprisingly, Belief-Space SIPS (BSIPS) is able to \emph{exactly} compute the posterior over beliefs, goals, and states (Equation \ref{eq:posterior}) without any Monte Carlo approximation: For the scenarios we considered, there were between 120 and 5940 possible combinations of goals, initial states, and beliefs. Enumerative inference over these hypotheses could run as fast as 0.1s/action (120 hypotheses), going up to 20s/action (5490 hypotheses). Experiments were conducted with a i7-1370P 1.90 GHz CPU and 64 GB RAM. Code can be found at \url{https://osf.io/xq9c4/}.

Algorithm \ref{alg:bsips} provides the pseudocode for BSIPS. At each step $t$, we simulate how the environment changes, and how the agent updates their beliefs based on what they see (L6-7). Next, we efficiently compute belief-space $Q$-values by leveraging the $Q_\text{MDP}$ approximation described in Section \ref{sec:modeling-perception-and-action}. This involves averaging over $Q$-values for each state hypothesis $\tilde s$ in the agent's belief $b_t^i$, which can be done cheaply by memoizing and reusing shortest-path computations across all belief hypotheses (L9-13). The $Q$-values allow us to compute the likelihood of the observed action $a_t$, allowing us to reweight each hypothesis by how well it explains the observations (L14-16)

For epistemic language evaluation at this scale, the technical challenges are mostly representational, not algorithmic. However, scaling to larger spaces of goals \cite{zhixuan2024infinite} and (belief) states will require additional layers of Monte Carlo approximation. Our implementation in Gen \cite{cusumano2019gen} can naturally be extended to these cases, e.g. by leveraging Sequential Monte Carlo for approximate inference of initial states \cite{del2006sequential,lew2023smcp3}.

\begin{algorithm}[H]
\footnotesize
\caption{Belief-Space Sequential Inverse Plan Search (BSIPS)}
\label{alg:bsips}
\begin{algorithmic}[1]
\Procedure{BSIPS}{$\mathcal{G}, \mathcal{S}_0, \mathcal{B}_0, a_{1:T}, o_{0:T}$}
    \State $\mathcal{H} \gets \mathcal{G} \times \mathcal{S}_0 \times \mathcal{B}_0$
    \Comment{Enumerate all hypotheses (goal, belief \& state combinations).}
    \State $\mathcal{W} \gets \{w^i := P(o_0 | s^i_0)\}_{i=1}^{|\mathcal{H}|}$
    \Comment{Initialize (unnormalized) weights for all hypotheses.}
    \For{$t \in [1, T]$}
        \For{$h^i := (g^i, s_{0:t-1}^i, b_{0:t-1}^i) \in \mathcal{H}$}
            \State $s_t^i \gets \textsc{state-transition}(s_{t-1}^i, a_{t-1})$
            \Comment{Simulate next environment state.}
            \State $b_t^i \gets \textsc{belief-update}(b_{t-1}^i, s_t^i, a_{t-1})$
            \Comment{Simulate agent's belief update.}
            \State $h^i \gets (g^i, s_{0:t}^i, b_{0:t}^i)$
            \State $Q_\text{Bel}(g^i, b_t^i, \tilde a) \gets 0$  \textbf{for} $\tilde a \in \textsc{valid-actions}(b_t^i)$
            \Comment{Initialize belief-space $Q$-values.}
            \For{$(\tilde s, \tilde w) \in b_t^i$ \textbf{and} $\tilde a \in \textsc{valid-actions}(\tilde s)$}
            \Comment{Iterate over environment states in agent's belief.}
                \State $Q^*(g^i, \tilde s, \tilde a) \gets \textsc{memoized}(\textsc{path-cost}(\tilde s, \tilde a, g^i))$
                \Comment{Compute shortest path cost to goal (memoized).}
                \State $Q_\text{Bel}(g^i, b_t^i, \tilde a) \gets Q^*(g^i, b_t^i, \tilde a) + \tilde w \cdot Q(g^i, \tilde s, \tilde a)$
                \Comment{Update belief-space $Q$-values.}
            \EndFor
            \State $P(a_t | b_t^i, g^i) \gets \exp(-\beta Q_\text{Bel}(g^i, b_t^i, a_t)) / \sum_{a} \exp(-\beta Q_\text{Bel}(g^i, b_t^i, a))$
            \Comment{Compute likelihood of action $a_t$.}
            \State $w^i \gets w^i \cdot P(a_t | b_t^i, g^i)$
            \Comment{Update weight with action likelihood.}
            \State $w^i \gets w^i \cdot P(o_t|s^i_t)$
            \Comment{Update weight with observation likelihood.}
        \EndFor
    \EndFor
    \State \Return $(\mathcal{H}, \mathcal{W})$
    \Comment{Return all hypotheses and their (unnormalized) weights.}
\EndProcedure
\end{algorithmic}
\end{algorithm}

\subsection{Parameter Fitting and Robustness Analyses}

We performed a grid search over parameters for the LaBToM model. The range of inverse temperatures $\beta$ for the Boltzmann policy went from $0.5$ to $4$ in multiplicative increments of $\sqrt{2}$. This produced human correlations between $r=0.75$ (at $\beta = 0.5$) and $r=0.81$ (at $\beta = 4$) for current beliefs, and between $r = 0.64$ (at $\beta = 0.5$) and $r=0.80$ (at $\beta = 2^{3/2}$), with $\beta = 2^{3/2}$ producing the best fit overall. Across all values of $\beta$, we found that the full model outperformed the ablated baselines.

We also fitted the threshold parameters $\Theta$ used in our ELoT representation. We performed grid-based coordinate ascent with a step size of 0.05, starting from values derived from the literature mapping modal words to probabilities \cite{wesson2009verbal,hahn2014grounding,meder2022developmental}, and limiting the search to a range of 0.2 above and below this starting point. Our initial and final threshold parameters are shown in Table \ref{tab:threshold-fitting}. To evaluate threshold sensitivity, we also ran the same procedure to \emph{minimize} correlation with humans. This produced a value of $r = 0.71$, which was still much higher than the next best model (GPT-4o I+Na+FS, $r = 0.52$). Optimizing these thresholds for the True Belief and Non-Planning ablations led to maximal correlations of $r = 0.12$ and $r = 0.47$ respectively.

\begin{table}[h]
    \centering
    \footnotesize
    \begin{tabular}{@{}lllllllllll@{}}
    \toprule
            & \multicolumn{10}{c}{\textbf{Thresholds} $\theta$}                                                                                                                                                      \\
            & $\textsf{believes}$ & $\textsf{certain}$ & $\textsf{uncertain}$ & $\textsf{likely}$ & $\textsf{unlikely}$ & $\textsf{could}$ & $\textsf{might}$ & $\textsf{may}$ & $\textsf{should}$ & $\textsf{must}$ \\ \midrule
    Initial & 0.75                & 0.95               & 0.50                 & 0.60              & 0.40                & 0.20             & 0.20             & 0.30           & 0.80              & 0.95            \\
    Fitted  & 0.75                & 0.95               & 0.70                 & 0.70              & 0.40                & 0.20             & 0.20             & 0.30           & 0.80              & 0.95            \\ \bottomrule
    \end{tabular}
        \caption{ELoT probability thresholds before and after fitting.}
    \label{tab:threshold-fitting}
\end{table}

\subsection{ELoT Translation Prompt}

For both grammar-constrained SMC and unconstrained sampling , we prompted the LLM with 36 examples to translate natural language statements into ELoT formulas. ELoT formulas were represented in a Prolog-like syntax analogous to the mathematical syntax we show in Table \ref{tab:elot}. The prompt used is shown below. We only show 10 out of 36 examples due to space constraints, and provide the full set in our code release. For translation of initial belief statements, we use a separate prompt to first translate the sentence from past tense to present tense, then translate the present tense sentence to ELoT.
\vspace{6pt}

\lstset{
    basicstyle=\tt\scriptsize,
    frame = single,
    breaklines=true,
    breakindent=10pt
}

\begin{lstlisting}
Please translate the statement below into logical form. Here are some examples of statements and their translations:

Input: The player knows that box 2 and box 3 are empty.
Output: knows_that(player, formula(and(empty(box2), empty(box3))))
Input: The player knows the color of the keys in all of the boxes.
Output: forall(box(B), knows_about(player, color(C), exists(and(key(K), inside(K, B)), iscolor(K, C))))
Input: The player doesn't know that there is a blue key in box 2.
Output: not_knows_that(player, formula(exists(and(key(K), iscolor(K, red)), inside(K, box2))))
Input: The player is sure of the color of the key in box 4.
Output: certain_about(player, color(C), exists(and(key(K), inside(K, box4)), iscolor(K, C)))
Input: The player is uncertain about what's in box 2.
Output: uncertain_about(player, color(C), exists(and(key(K), inside(K, box2)), iscolor(K, C)))
Input: The player believes that there is a key in box 4.
Output: believes(player, formula(exists(key(K), inside(K, box4))))
Input: The player thinks that there is a red key in either box 1 or box 3.
Output: believes(player, formula(exists(and(key(K), iscolor(K, red)), or(inside(K, box1), inside(K, box3)))))
Input: The player thinks there might be a key in box 1 or box 2.
Output: believes(player, might(exists(key(K), or(inside(K, box1), inside(K, box2)))))
Input: The player thinks there is likely a key in box 2.
Output: believes(player, likely(exists(key(K), inside(K, box2))))
\end{lstlisting}

\subsection{LLM Baseline Prompts}

Below we show the prompts we used for our multimodal LLM baselines (GPT-4o and Gemini 1.5 Pro). Associated images and videos can be found in our code and dataset release.

\begin{lstlisting}
[IMAGE OR VIDEO]

You're watching someone play the treasure game shown above.
                  
The player controls a character, and their goal is collect one of the four gems (triangle, square, hexagon, or circle).

The rules of the game are as follows:
  - The player can move on the white squares.
  - The player has a full view of the map at all time.
  - The player's goal is to collect exactly one target gem.
  - Keys unlock doors of the same color (e.g. red keys unlock red doors).
  - Each key can only be used once. Keys disappear after use.
  - Each box may be empty or contain exactly one key.
  - The player may or may not know what's in each box.
  - Neither you nor the player can see what's hidden in each box. But both of you can see all other objects in the scene.
  - There are at most two keys hidden among the boxes.
  - The player knows that the puzzle is solvable, which means there are just enough keys to reach any of the target gems.
  - The keys and doors are labeled. The labels are shown on the top right corner of each cell.
  - Your task is to figure out what the player's goal is, and also what the player initially believed about the contents of the boxes.

Now you observe the following:

[INSERT PLAN OR NARRATIVE]

Given this information, which gem(s) are most likely to be the human agent's goal? And how would you rate the following statement about the player's current belief from 1 (definitely false) to 7 (definitely true)? Rate 4 if you think there is an equal chance of the statement being true and false.

Please rate the following statement:

[INSERT STATEMENT]

Please respond in the following JSON format, indicating all gems that you think are likely to be the human's goal, and your rating as a number from 1 to 7. 

{
goal: [gems...],
rating: x ,
}

The gems should be any of [triangle, square, hexagon, circle] and you can indicate all the likely goal gems in your response. The rating should be an integer from 1 to 7. Please provide an explanation to your response.
\end{lstlisting}
An example of plan-based prompting:
\begin{lstlisting}
The player moves right three times, then down twice, and finally left.
\end{lstlisting}
An example of narrative-based prompting, which provides more contextual information:
\begin{lstlisting}
The square gem is locked behind a red door. The triangle, circle and hexagon gems are not locked behind any doors. There are three boxes. No keys are visible in the scene. 
The player moves right three times, then down twice, and finally left towards box 2 and away from box 3.
\end{lstlisting}
For few shot prompting, the following text is added before showing the statement to be evaluated. We provide 4 examples of human ratings:
\begin{lstlisting}
Here is how other people have rated other statements for this stimulus:

{% for statement in statements %}
statement {{ loop.index }}: {{ statement.text }}
rating: {{ statement.rating }}

{% end for %}
\end{lstlisting}

\pagebreak

\section{Additional Results and Experiment Details}

\subsection{ELoT Parser Evaluation}

\begin{wraptable}[15]{R}{0.625\textwidth}
\centering
\ssmall
\vspace{-9pt}
\begin{tabular}{@{}rl@{}}
\toprule
\multicolumn{2}{c}{\textbf{LLaMa 3.1 8B + Grammar-Constrained SMC}}                                                                                                                                                   \\ \midrule
\textbf{Statement}   & The player initially expected to find a key in box 3.                                                                                                                                          \\
\textbf{Gold Trans.} & \textsf{believes(player, formula(exists(key(K), inside(K, box3))))}                                                                                                                            \\
\textbf{Translation} & \begin{tabular}[c]{@{}l@{}}(67\%) \textsf{believes(player, likely(exists(key(K), inside(K, box3))))}\\ (32\%) \textsf{believes(player, formula(exists(key(K), inside(K, box3))))}\end{tabular} \\
\textbf{Error}       & ``Expected'' is interpreted as ``likely'' instead of straightforward belief.                                                                                                                   \\ \midrule
\multicolumn{2}{c}{\textbf{LLaMa 3.1 8B Unconstrained + Majority Vote}}                                                                                                                                               \\ \midrule
\textbf{Statement}   & The player initially thought that box 2 contained a red key.                                                                                                                                   \\
\textbf{Gold Trans.} & \textsf{believes(player, formula(exists(and(key(K), iscolor(K, red)), inside(K, box2))))}                                                                                                      \\
\textbf{Translation} & \begin{tabular}[c]{@{}l@{}}(10\%) \textsf{believes(player, contains(red, box2))}\\ (10\%) \textsf{type about(player, box2, contains)}\end{tabular}                                             \\
\textbf{Error}       & Syntax errors, out of vocabulary term ``contains''.                                                                                                                                            \\ \midrule
\multicolumn{2}{c}{\textbf{Gemini Flash 8B Unconstrained + Majority Vote}}                                                                                                                                            \\ \midrule
\textbf{Statement}   & The player initially did not know if box 2 had a key.                                                                                                                                          \\
\textbf{Gold Trans.} & \textsf{not\_knows\_if(player, formula(exists(key(K), inside(K, box2))))}                                                                                                                      \\
\textbf{Translation} & (100\%) \textsf{not\_knows\_if(player, exists(key(K), inside(K, box2)))}                                                                                                                       \\
\textbf{Error}       & Missing ``formula'' predicate to denote base formula.                                                                                                                                          \\ \bottomrule
\end{tabular}
\vspace{-3pt}
\caption{Erroneous ELoT translations for different parsing methods.}
\label{tab:translation_errors}
\end{wraptable}

When evaluating translations, we performed both an exact equality check against our gold translations, and also a manual check for approximate semantic equivalence (i.e. cases where the parser gave a reasonable alternative translation of the original English statement). For replicability, we include these manual equivalence annotations in our code release.

Table \ref{tab:translation_errors} shows characteristic ELoT translation errors produced by each parsing method. Many of the the strict errors from grammar-constrained SMC pass the approximate equivalence check, and SMC often produces a distribution over reasonable alternative translations (Row 1). In contrast, unconstrained sampling from the same LLaMa 3.18B base model leads to many syntax errors that majority voting cannot correct for (Row 2). With Gemini Flash 8B, we observe the mode collapse issue that frequently afflicts finetuned LLMs \cite{o'mahony2024attributing}, often leading all $n_\sigma$=$10$ samples to produce the same incorrect translation (Row 3).

\subsection{Unimodal LLM Baseline with Symbolic PDDL Inputs}

\begin{wraptable}[7]{R}{0.5\textwidth}
    \centering
    \footnotesize
    \vspace{-6pt}
    \begin{tabular}{ccc}
        \toprule
        \textbf{Prompt Format} & \textbf{Current Beliefs} & \textbf{Initial Beliefs} \\
        \midrule
        Image + Narrative + FS & 0.52 (0.01) & 0.41 (0.01)\\
        PDDL + Narrative + FS & 0.32 (0.01) & 0.20 (0.01)\\
        \bottomrule
    \end{tabular}
    \caption{GPT-4o correlation with human ratings across different input representations.}
    \label{tab:llm_input}
\end{wraptable}

In our experiment, we used raw visual inputs, such as images or videos, for the multimodal LLM baselines. As the LaBToM model operates over symbolic representations of agent and environment states encoded in the Planning Domain Definition Language (PDDL), we also experimented with prompting LLMs with symbolic PDDL inputs. We report the results on the best performing LLM baseline (GPT4o I+Na+FS) in Table \ref{tab:llm_input}, which shows that replacing raw visual input with PDDL (as text) hinders performance. 

\subsection{Impact of a Normalized Statement Prior}

In the main text, we report results under the assumption that human observers respond as if they have a normalized 50-50 prior $U_{\varphi}$ over whether each statement $\varphi$ is true. Under this assumption, the posterior truth-value of a statement $P(\llbracket \varphi \rrbracket_{(s_t, b_t)}| a_{1:T}, o_{0:T})$ can be interpreted as a \emph{normalized likelihood}:
\begin{equation}
\bar L(\llbracket \varphi \rrbracket_{(s_t, b_t)}| a_{1:T}, o_{0:T}) = \frac{P(a_{1:T}, o_{0:T} | \llbracket \varphi \rrbracket_{(s_t, b_t)})}{P(a_{1:T}, o_{0:T} | \llbracket \varphi \rrbracket_{(s_t, b_t)}) + P(a_{1:T}, o_{0:T} | \llbracket \neg\varphi \rrbracket_{(s_t, b_t)})}
\end{equation}
An alternative assumption is to use a uniform prior $U_{\mathcal{S}_0 \times \mathcal{B}_0}$ over all possible initial states $\mathcal{S}_0$ and belief distributions $\mathcal{B}_0$. This has the effect of up-weighting statements which are true in \emph{more possible worlds}, e.g. \emph{``The player believes that a red key might be in box 1, 2, or 3.''}.

Table \ref{tab:normalized_vs_unnormalized} shows the impact of making either assumption, in terms of the Pearson's correlation coefficient (PCC) $r$ and mean absolute error (MAE) with human judgments, assuming the model has the gold translations. Consistent with \citet{ying2024grounding}, using a normalized statement prior largely improves the correlation while reducing the mean absolute error for the full LaBToM model. In other words, people appear more willing to say that a statement $\varphi$ is true only if they have \emph{evidence} for $\varphi$, and otherwise default to a 50-50 rating. We see sharper differences for initial belief statements, which is likely because priors have a stronger effect on initial beliefs, whereas an agent's current beliefs are more strongly determined by their percepts: If an agent sees that a box is empty, an observer's judgment about whether the agent believes that the box is empty should not depend on the observer's prior.

\begin{table}[t]
\centering
\footnotesize
\begin{tabular}{@{}llllllll@{}}
\toprule
\multicolumn{1}{c}{\textbf{Model}} & \multicolumn{1}{c}{\textbf{Prior}}       & \multicolumn{2}{c}{\textbf{Overall}}                  & \multicolumn{2}{c}{\textbf{Current}}                  & \multicolumn{2}{c}{\textbf{Initial}}                  \\
\multicolumn{1}{c}{}               &                                          & \multicolumn{1}{c}{PCC $r$ $\uparrow$} & \multicolumn{1}{c}{MAE $\downarrow$} & \multicolumn{1}{c}{PCC $r$ $\uparrow$} & \multicolumn{1}{c}{MAE $\downarrow$} & \multicolumn{1}{c}{PCC $r$ $\uparrow$} & \multicolumn{1}{c}{MAE $\downarrow$} \\ \midrule
LaBToM (ours)                      & $U_\varphi$                              & \textbf{0.81 (0.01)}        & \textbf{0.195 (0.004)}  & 0.81 (0.01)                 & \textbf{0.193 (0.006)}  & \textbf{0.80 (0.01)}        & \textbf{0.196 (0.006)}  \\
                                   & $U_{\mathcal{S}_0 \times \mathcal{B}_0}$ & 0.78 (0.01)                 & 0.226 (0.004)           & \textbf{0.82 (0.01)}        & 0.211 (0.006)           & 0.71 (0.01)                 & 0.244 (0.005)           \\
Non-Planning                       & $U_\varphi$                              & 0.47 (0.01)                 & 0.200 (0.003)           & 0.60 (0.02)                 & 0.185 (0.004)           & 0.20 (0.01)                 & 0.219 (0.006)           \\
                                   & $U_{\mathcal{S}_0 \times \mathcal{B}_0}$ & 0.55  (0.01)                & 0.252 (0.003)           & 0.70 (0.01)                 & 0.215 (0.005)           & 0.34 (0.02)                 & 0.300 (0.005)           \\
True Belief                        & $U_\varphi$                              & 0.12 (0.01)                 & 0.475 (0.003)           & 0.09 (0.01)                 & 0.482 (0.004)           & 0.13 (0.02)                 & 0.466 (0.004)           \\
                                   & $U_{\mathcal{S}_0 \times \mathcal{B}_0}$ & 0.12 (0.01)                 & 0.475 (0.003)           & 0.09 (0.01)                 & 0.482 (0.004)           & 0.13 (0.02)                 & 0.467 (0.004)           \\ \bottomrule
\end{tabular}
\caption{Similarity of human and model ratings for a normalized ($U_\varphi$) vs. unnormalized ($U_{\mathcal{S}_0 \times \mathcal{B}_0}$) prior.}
\label{tab:normalized_vs_unnormalized}
\end{table}

\subsection{Per-Factor Model Performance using Gold Translations}

Table \ref{tab:correlation_breakdown} shows the correlation between human judgments and model outputs when using the gold ELoT translations. Performance is broken down by the annotated factors described in Table \ref{tab:dataset_overview}. Using gold ELoT translations improves LaBToM's correlation with human judgments from $r$=$0.76$ (Table \ref{tab:performance_summary}) to $r$=$0.81$.

Additionally, LaBToM robustly outperforms the baselines on almost all of these data splits, achieving a correlation around $r$=$0.8$ in each case. The one exception is the set of statements about what the agent initially knows (\emph{Init. Know.}, $r=0.31$), such as \emph{``The player initially knew that the red key was in box 3''}. As we discuss in the next section, this is likely because human participants assume that \emph{direct perception} or \emph{justification} is necessary for other agents to know some proposition $\phi$. In contrast, our model treats knowledge claims as equivalent to claims of true belief.

\begin{table}[h!]
\centering
\footnotesize
\begin{tabular}{@{}lllllll@{}}
\toprule
\multicolumn{2}{c}{\textbf{Model}} & \multicolumn{5}{c}{\textbf{Human Correlation $r$} (s.e.)}                                                               \\
                  &                & \textit{Current}     & \textit{Curr. Poss.} & \textit{Curr. Prob.} & \textit{Curr. Comp.} & \textit{Curr. Know.} \\ \midrule
LaBToM            & Full (ours)    & \textbf{0.81 (0.01)} & \textbf{0.80 (0.02)} & \textbf{0.76 (0.03)} & \textbf{0.76 (0.03)} & \textbf{0.89 (0.01)} \\
                  & Non-Planning   & 0.60 (0.02)          & 0.65 (0.02)          & 0.48 (0.04)          & 0.34 (0.04)          & 0.75 (0.02)          \\
                  & True Belief    & 0.09 (0.01)          & 0.14 (0.03)          & 0.17 (0.03)          & 0.02 (0.03)          & 0.43 (0.02)          \\ \midrule
GPT-4o            & I, Na, FS      & 0.59 (0.01)          & 0.62 (0.03)          & 0.63 (0.03)          & 0.64 (0.02)          & 0.67 (0.02)          \\
                  & I, Na          & 0.52 (0.01)          & 0.50 (0.03)          & 0.54 (0.04)          & 0.61 (0.03)          & 0.58 (0.02)          \\
                  & I, Pl          & 0.32 (0.01)          & 0.50 (0.03)          & 0.30 (0.03)          & 0.42 (0.02)          & 0.21 (0.02)          \\ \midrule
Gemini 1.5 Pro    & V, Na, FS      & 0.28 (0.01)          & 0.23 (0.03)          & 0.32 (0.03)          & 0.10 (0.03)          & 0.28 (0.02)          \\
                  & I, Na, FS      & 0.29 (0.01)          & 0.25 (0.03)          & 0.39 (0.03)          & 0.17 (0.03)          & 0.26 (0.02)          \\ \midrule
                  &                & \textit{Initial}     & \textit{Init. Poss.} & \textit{Init. Prob.} & \textit{Init. Comp.} & \textit{Init. Know.} \\ \midrule
LaBToM            & Full (ours)    & \textbf{0.80 (0.01)} & \textbf{0.84 (0.02)} & \textbf{0.81 (0.03)} & \textbf{0.78 (0.02)} & 0.31 (0.05)          \\
                  & Non-Planning   & 0.20 (0.01)          & 0.32 (0.02)          & -0.05 (0.03)         & 0.21 (0.03)          & 0.34 (0.05)          \\
                  & True Belief    & 0.13 (0.02)          & -0.09 (0.03)         & 0.12 (0.05)          & -0.03 (0.04)         & 0.31 (0.05)          \\ \midrule
GPT-4o            & I, Na, FS      & 0.41 (0.01)          & 0.41 (0.03)          & 0.01 (0.03)          & 0.26 (0.03)          & 0.31 (0.05)          \\
                  & I, Na          & 0.41 (0.01)          & 0.30 (0.03)          & 0.48 (0.04)          & 0.31 (0.02)          & \textbf{0.51 (0.05)} \\
                  & I, Pl          & 0.18 (0.01)          & 0.05 (0.03)          & 0.15 (0.04)          & 0.23 (0.03)          & 0.30 (0.05)          \\ \midrule
Gemini 1.5 Pro    & V, Na, FS      & 0.14 (0.02)          & -0.02 (0.03)         & -0.04 (0.04)         & -0.08 (0.03)         & 0.27 (0.05)          \\
                  & I, Na, FS      & 0.11 (0.02)          & 0.02 (0.03)          & -0.04 (0.04)         & -0.02 (0.03)         & 0.29 (0.05)          \\ \bottomrule
\end{tabular}
\caption{Human vs. model correlations broken down by linguistic factors. LaBToM results use gold translations.}
\label{tab:correlation_breakdown}
\end{table}

\pagebreak

\subsection{Differences in Human Ratings vs. Model Inferences}

While LaBToM largely matches human evaluations of epistemic language both in aggregate and at the individual scenario level, there a number of interesting ways in which they differ.

\textbf{People are less certain than LaBToM.} One difference is simply that our model tends to be more certain than people, using the extremes of the 0-1 scale in ways that our participants tended to avoid. This effect did not appear to be driven by the choice of the Boltzmann inverse temperature $\beta$, since lower values of $\beta$ (which increase model uncertainty) led to poorer fits with human data. Instead, humans may be evaluating the truth a statement $\phi$ less strictly than our model does, perhaps by maintaining uncertainty over the probability thresholds $\theta$ associated with each statement.

\textbf{People appear to adapt probability thresholds.} Threshold uncertainty is closely related to another potential driver of difference: Unlike our model, participants appear to contextually adapt probability thresholds associated with modal words, in line with work on the pragmatics of epistemic modals \cite{schuster2020know,rudin2016deriving,lassiter2017graded}. This is evinced by human responses for current belief statement S2 in Fig. \ref{fig:qualitative}b. People rate \emph{``The player is unsure which box has a key''} highly despite the apparent confidence that the player exhibits in looking for a key in box 3 (at the expense of looking in box 1 or box 2). This is consistent with an upwards adjustment of $\theta_\textsf{uncertain}$ from 0.55, such that the player is judged as uncertain even when they seem to think it is quite likely for a key to be in box 3. In contrast, our model thinks it is unlikely that the player is uncertain. Similar effects can be seen for current statement S1 and S5 in the same scenario, except that people appear to adjust their threshold for \emph{may} and \emph{might} downwards from $0.3$ and $0.2$ to accommodate even highly unlikely possibilities.

\begin{figure}[t]
    \centering
    \includegraphics[width=\linewidth]{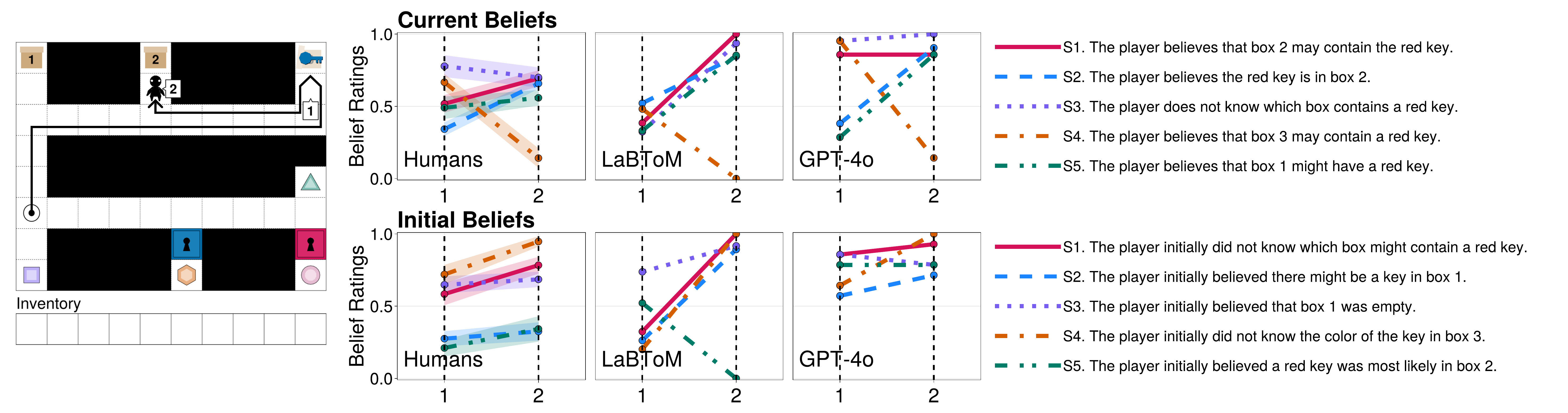}
    \caption{Scenario illustrating differences in human and model ratings of knowledge claims.}
    \label{fig:storyboard_p3_2}
    \vspace{-9pt}
\end{figure}

\textbf{People respond as if knowledge requires justification.} Unlike our model, which reduces knowledge statements to true belief claims (Table \ref{tab:elot}), people appear to assume that some form of justification (e.g. via direct perception) is necessary for agents to \emph{know} that some proposition $\phi$ is true. This is clearly illustrated in Figure \ref{fig:storyboard_p3_2}: In this scenario, the player very confidently walks past boxes 1 and 2 towards box 3, suggesting a strong belief about its contents. Since our model just treats knowledge as true belief (and it is quite possible that the player is correct), LaBToM assigns less than 50\% probability to \emph{``The player does not know which box contains a red key''} and \emph{``The player initially did not know the color of the key in box 3''} at Judgment Point 1 (which occurs \emph{before} box 3 is opened). Human judgments differ significantly, assigning more than 50\% to these ignorance claims. This is consistent with people understanding knowledge to require justification or direct perception: Since the player has not seen what is in box 3 by Judgment Point 1,  they cannot \emph{know} what is in that box.

\textbf{People reason boundedly about logical implications.} A final difference between humans and our model is that people appear to exhibit bounded reasoning when evaluating statements with non-obvious implications. In Figure \ref{fig:storyboard_p3_2}, for example, \emph{``The player initially believed a red key was most likely in box 2''} is rated lowly by humans initially, then more highly, whereas our model assigns a zero rating to that statement by Judgment Point 2. This is because by then, it is clear that the player did not need a blue key, and was instead looking for a red key in box 1. The fact that they looked in box 1 first also implies that they initially thought box 1 was most likely to contain a red key, not box 2. Our model captures this reasoning, but people do not seem to independently grasp these multi-step implications, consistent with studies on the boundedness of human reasoning \cite{smets2018effort,mercier2017enigma}.

\end{document}